\documentclass[sigconf]{acmart}
\renewcommand\footnotetextcopyrightpermission[1]{} 
\AtBeginDocument{%
	}

\usepackage{multirow}
\usepackage{booktabs}
\usepackage{makecell}
\usepackage{colortbl}
\usepackage{pifont}
\usepackage{bbm}

\begin{document}
	
\title{Seeing Like Radiologists: Context- and Gaze-Guided Vision-Language Pretraining for Chest X-rays }

\author{Kang Liu}
\affiliation{%
\institution{Xidian University}
\city{Xi'an}
\country{China}
}
\email{kangliu@stu.xidian.edu.cn}

\author{Zhuoqi Ma}
\affiliation{%
	\institution{Xidian University}
	\city{Xi'an}
	\country{China}
}
\email{zhuoqima@xidian.edu.cn}

\author{Siyu Liang}
\affiliation{%
	\institution{Xidian University}
	\city{Xi'an}
	\country{China}
}
\email{syliang\_233@stu.xidian.edu.cn}

\author{Yunan Li}
\affiliation{%
	\institution{Xidian University}
	\city{Xi'an}
	\country{China}
}
\email{yunanli@xidian.edu.cn}

\author{Xiyue Gao}
\affiliation{%
	\institution{Xidian University}
	\city{Xi'an}
	\country{China}
}
\email{xygao@xidian.edu.cn}

\author{Chao Liang}
\affiliation{%
	\institution{Wuhan University}
	\city{Wuhan}
	\country{China}
}
\email{cliang@whu.edu.cn}

\author{Kun Xie}
\authornote{Corresponding authors.}
\affiliation{%
	\institution{Xidian University}
	\city{Xi'an}
	\country{China}
}
\email{xiekun@xidian.edu.cn}

\author{Qiguang Miao}
\authornotemark[1]
\affiliation{%
	\institution{Xidian University}
	\city{Xi'an}
	\country{China}
}
\email{qgmiao@xidian.edu.cn}


\begin{abstract}
Despite recent advances in medical vision-language pretraining, existing models still struggle to capture the diagnostic workflow: radiographs are typically treated as context-agnostic images, while radiologists' gaze---a crucial cue for visual reasoning---remains largely underexplored by existing methods. These limitations hinder the modeling of disease-specific patterns and weaken cross-modal alignment. To bridge this gap, we introduce \textbf{CoGaze}, a \textbf{Co}ntext- and \textbf{Gaze}-guided vision-language pretraining framework for chest X-rays. We first propose a context-infused vision encoder that models how radiologists integrate clinical context---including patient history, symptoms, and diagnostic intent---to guide diagnostic reasoning. We then present a multi-level supervision paradigm that (1) enforces intra- and inter-modal semantic alignment through hybrid-positive contrastive learning, (2) injects diagnostic priors via disease-aware cross-modal representation learning, and (3) leverages radiologists' gaze as probabilistic priors to guide attention toward diagnostically salient regions. Extensive experiments demonstrate that CoGaze consistently outperforms state-of-the-art methods across diverse tasks, achieving up to \textbf{+2.0\% CheXbertF1} and \textbf{+1.2\% BLEU2} for free-text and structured report generation, \textbf{+23.2\% AUROC} for zero-shot classification, and \textbf{+12.2\% Precision@1} for image-text retrieval. Code is available at \href{https://github.com/mk-runner/CoGaze}{https://github.com/mk-runner/CoGaze}.
\end{abstract}

%



\maketitle

\section{Introduction}
Vision-language pretraining (VLP) has emerged as a powerful paradigm for learning generalizable and transferable multimodal representations, driven by the rise of large-scale datasets and multimodal supervision \cite{lecun2015deep,tpami-foundation-survey,2025-foundation-model-medicine,ma2025fully-ark-Nature}. In the natural image domain, models such as CLIP \cite{radford-learning-clip} and BLIP \cite{pmlr-blip,li2023-blip2} have achieved remarkable cross-modal alignment, inspiring efforts to extend VLP to medical imaging \cite{zhang2025biomedclipmultimodalbiomedicalfoundation,huang2024-NC-MaCo,zhou2023-iclr-MRM,medclip_wang_2022}. These methods leverage paired or unpaired image-report data to learn task-agnostic representations, offering a unified backbone for diverse downstream tasks.

\begin{figure}
\centering
\includegraphics[width=1\linewidth]{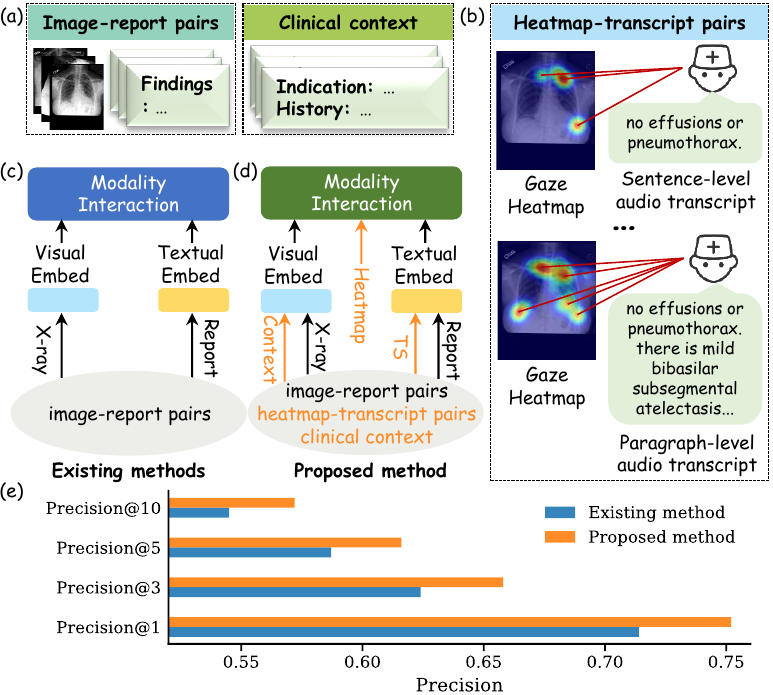}
\caption{Overview of the pretraining framework. (a) Image-report pairs with optional clinical context. (b) Heatmap-transcript pairs capturing physicians' visual attention. (c-d) Comparison between existing methods and ours. (e) Image-text retrieval results on the MIMIC-5x200 dataset \cite{zhou2024benchx}. Clinical context and gaze data are available for a subset of cases. ``TS'' denotes transcript.}
\label{fig:1-motivation}
\end{figure}

Despite these advances, directly transferring natural-image VLP strategies to medical imaging remains challenging due to the limited dataset scale and high cost of expert annotation. Existing medical VLP frameworks \cite{zhou2022-NAI-REFERS,cheng-prior} rely primarily on chest X-ray image-report pairs (Fig.~\ref{fig:1-motivation}(c)) and introduce auxiliary objectives to alleviate data scarcity. For instance, MGCA \cite{wang-mgca} maximizes cross-modal correspondence via multi-granularity alignment, MRM \cite{zhou2023-iclr-MRM} reconstructs masked patches for fine-grained semantic understanding, and KAD \cite{zhang-kad} infuses domain knowledge to improve reasoning. However, these approaches typically treat radiographs as context-agnostic inputs, overlooking critical clinical priors (i.e., patient symptoms and medical history) and underexploring radiologists' gaze, which provides valuable cues about diagnostic focus. Consequently, while they achieve strong image-report alignment, they fail to capture the reasoning process underlying radiological diagnosis, resulting in representations that lack clinical grounding and practical utility.

Recent studies in chest X-ray report generation demonstrate that incorporating clinical context---including patient symptoms and medical history---yields more accurate and clinically coherent reports \cite{ml4h-indication-rg,sei,bannur2024maira2groundedradiologyreport,acl-2025-libra}. Nevertheless, most existing medical VLP frameworks \cite{2509-generative-foundation-model,2025-tmi-foundation-model-bbox,2024-eva-x-foundation-model,2025-wacv-foundation-x} still treat chest X-rays as context-agnostic images. To address this gap, we explicitly encode clinical context into the pretraining process, aligning representation learning with real-world diagnostic reasoning.

Radiologists' gaze offers a promising yet underexplored supervision source, revealing diagnostic focus and spatial attention patterns (Fig.~\ref{fig:1-motivation}(b)). Prior work has demonstrated that incorporating gaze information can enhance performance in disease classification \cite{2024-accv-eye-cls,riju2025-arxiv-eye} and report generation \cite{acl-eye-2025-look,2024-ACCV-eye-gaze-fg-cxr}, suggesting its potential for learning semantically rich visual representations. However, methods for integrating gaze signals into medical VLP are still in their infancy. For instance, \citeauthor{2024-miccai-eye-vlm} overlays gaze heatmaps onto images, introducing mixed visual signals that may be misinterpreted as image content rather than attention guidance. EGMA \cite{ma2024-nips-eye-egma} converts gaze coordinates into binary masks for multimodal alignment, but this binarization oversimplifies gaze supervision and neglects the continuous nature of spatial attention---where fixation points should carry higher importance with smoothly decaying influence in surrounding areas. These limitations motivate our framework, which models gaze as a soft probabilistic prior, enabling fine-grained and continuous modeling of spatial attention.

In this paper, we introduce \textbf{CoGaze}, a \textbf{Co}ntext- and \textbf{Gaze}-guided vision-language pretraining framework for chest X-rays. We first present a context-infused vision encoder that jointly encodes view position, clinical context, and visual semantics within a unified representation space, mirroring the clinical workflow in which radiologists interpret images guided by patient information and diagnostic intent (i.e., clinical context). To further enhance representation learning, we propose a multi-level supervision paradigm that enforces clinically grounded alignment across different granularities: (1) global alignment via hybrid-positive contrastive learning, which unifies single- and multi-positive contrastive learning to achieve both intra- and inter-modal semantic alignment; (2) disease-aware cross-modal representation learning, which aligns images and reports within a shared disease label space to enrich visual features with diagnostic priors; and (3) fine-grained attention via soft gaze guidance, which treats radiologists' gaze as probabilistic priors to couple salient image regions with corresponding textual descriptions, embedding diagnostic attention into the representation space. Extensive experiments across diverse downstream tasks demonstrate the effectiveness of CoGaze. Our contributions are:
\begin{itemize}
\item We present CoGaze, a clinically grounded vision-language pretraining framework for chest X-rays that integrates view position, clinical context, and radiologists' gaze into a unified representation learning pipeline, reflecting real-world diagnostic reasoning.
\item We propose a multi-level supervision paradigm that combines (i) hybrid-positive contrastive learning for global semantic alignment, (ii) disease-aware cross-modal classification for diagnostic prior infusion, and (iii) soft gaze guidance for fine-grained attention modeling.
\item Extensive experiments demonstrate that CoGaze consistently outperforms state-of-the-art baselines, achieving up to +2.0\% CheXbertF1 and +1.2\% BLEU2 for free-text and structured report generation, +23.2\% AUROC for zero-shot classification, and +12.2\% Precision@1 for retrieval. 
\end{itemize}

\section{Related Work}

\textbf{Chest X-ray Vision-Language Models.}  Vision-language models (VLMs) have shown strong potential for generalizable medical image understanding, yet their application to chest X-rays remains constrained by the limited scale of paired image-report data. To address this issue, MedCLIP \cite{medclip_wang_2022} introduces a semantic matching loss to exploit unpaired datasets, REFERS \cite{zhou2022-NAI-REFERS} enforces multi-view consistency across studies, and MaCo \cite{huang2024-NC-MaCo} applies masked contrastive learning for fine-grained representation learning. Recent large-scale medical VLMs such as Med-PaLM \cite{2024-nejm-ai-med-palm}, CheXagent \cite{chen2024-chexagent}, LLaVA-Med \cite{NEURIPS2023-llava-med}, and LLaVA-Rad \cite{zambrano2025-llava-rad} adapt general-purpose vision-language architectures to the medical domain through prompt-based reasoning, achieving improved performance across diverse downstream tasks. However, existing models rely primarily on static image-report alignment and overlook key components of diagnostic reasoning---such as clinical context and radiologists' visual attention---that are essential for clinically meaningful representation learning. To bridge this gap, we introduce CoGaze, a context- and gaze-guided vision-language pretraining framework that explicitly encodes clinical context and incorporates gaze-informed supervision to enhance alignment with the diagnostic workflow.

\begin{figure*}
\centering
\includegraphics[width=1\linewidth]{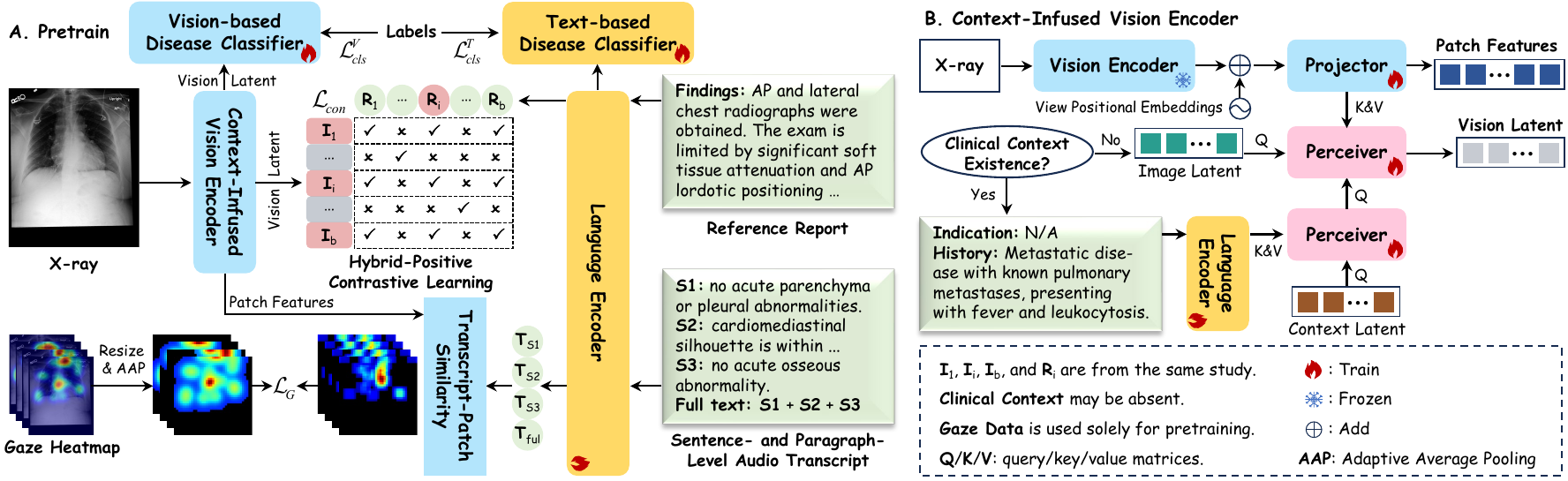}
\caption{(A) Overview of our CoGaze. (B) Context-infused vision encoder. Gaze supervision is used only during pretraining.}
\label{fig:2-overview}
\end{figure*}

\textbf{Eye-tracking for Modeling Diagnostic Attention.} Eye-tracking datasets, such as EGD \cite{2020-eye-gaze-data} and REFLACX \cite{2022-reflacx}, record radiologists' gaze trajectories along with synchronized spoken transcripts, providing fine-grained supervision for modeling visual attention and diagnostic reasoning (Fig.~\ref{fig:1-motivation}(b)). Prior studies utilize these signals in two forms: heatmap-based encodings \cite{2024-miccai-eye-vlm,ma2024-nips-eye-egma,2024-ACCV-eye-gaze-fg-cxr} and textual prompts \cite{wang2024-wacv-gazegnn,acl-eye-2025-look}. Textual prompts offer semantic interpretability but lack spatial specificity, whereas heatmaps retain pixel-level attention patterns that more directly reflect diagnostic focus. Heatmap-based approaches have shown that incorporating gaze as expert supervision improves visual representations for specific tasks, such as disease classification \cite{2024-accv-eye-cls,riju2025-arxiv-eye} and report generation \cite{2024-ACCV-eye-gaze-fg-cxr,acl-eye-2025-look}. However, these efforts are limited to isolated tasks, and their insights have yet to be fully explored in medical VLP. This gap motivates frameworks that leverage gaze as a supervisory signal to learn generalizable, clinically grounded representations.

\section{Method}
\subsection{Problem Formulation}
Fig.~\ref{fig:2-overview}(A) provides an overview of CoGaze, a context- and gaze-guided vision-language pretraining framework for chest X-rays. The objective is to learn clinically grounded visual representations that are transferable to a variety of downstream tasks. Formally, given a chest X-ray image $x_i$ with optional clinical context $c_i$ and gaze annotations $g_i$, the encoder $f_{\theta}$ maps these inputs into a latent representation ${\boldsymbol{Z}_i}=f_{\theta}(x_i,c_i,g_i)$. Here, $c_i$ and $g_i$ are available only for a subset of samples, with $g_i$ used exclusively during pretraining. The resulting representation integrates visual semantics, contextual information, and radiologists' gaze, forming a unified feature space that supports diverse downstream tasks, including report generation, classification, segmentation, and image-text retrieval.

\subsection{Dual Encoders for Vision and Language}
\textbf{Shared Language Encoder.} Building on advances in language modeling \cite{devlin-etal-2019-bert,PriorRG}, we adopt a unified language encoder to process heterogeneous clinical texts. Task-specific special tokens (i.e., [Findings], [Transcript], [Indication], [History]) are prepended to enable role-aware representations with minimal parameter overhead. For efficiency, \textit{indication} and \textit{history} sections are concatenated into a single clinical context sequence: [Indication]\{\textit{indication}\}[History] \{\textit{history}\}). The resulting sequence is encoded into contextual embeddings ${\boldsymbol{C}}_i \in {\mathbb R}^{n_c \times d}$, where $n_c$ is the token length and $d$ the embedding dimension. Similarly, reference reports are encoded as $\boldsymbol{R}_i \in {\mathbb R}^{n_r \times d}$. Audio transcripts are processed hierarchically into sentence- and paragraph-level embeddings ${\boldsymbol{T}_i} = \{\boldsymbol{T}_\text{S1},\boldsymbol{T}_\text{S2},\dots,\boldsymbol{T}_\text{full}\} \in {\mathbb R}^{n \times s \times d}$, where $n$ denotes the number of segments and $s$ the maximum token length per segment.

\textbf{Context-Infused Vision Encoder.} In clinical practice, radiologists interpret chest X-rays by integrating imaging evidence with contextual information (i.e., \textit{Indication} and \textit{History}) to support diagnostic reasoning. Inspired by this process, we propose a context-infused vision encoder (Fig.~\ref{fig:2-overview}(B)) that models view position and clinical context to enrich visual representations. Patch-level features are first extracted from the input X-ray and augmented with learnable view-positional embeddings, where an ``unknown'' embedding is assigned to unspecified views. The features are then projected into patch features ${\boldsymbol{X}_i^p} \in {\mathbb{R}^{p \times d}}$, where $p$ denotes the number of patches and $d$ the feature dimension. 

To robustly address missing clinical context, we introduce a context-adaptive encoding mechanism. When context is available, context embeddings ${\boldsymbol{C}}_i$ from the language encoder are fused with learnable context latents $\boldsymbol{Z}_C \in {\mathbb{R}^{m \times d}}$ through a Perceiver \cite{icml-2021-perceiver} module, yielding compressed context features $\bar{\boldsymbol{C}}_i \in {\mathbb{R}^{m \times d}}$. Here $m \ll p$ is the number of latents. If context is absent, a dedicated learnable image latent $\boldsymbol{Z}_I \in {\mathbb{R}^{m \times d}}$ serves as a surrogate. Formally,
\begin{align}
\bar{\boldsymbol{C}}_i = \begin{cases}
\text{Perceiver}(\boldsymbol{Z}_C,\boldsymbol{C}_i)   & \text{if    } \boldsymbol{C}_i \text{    exists},  \\
\boldsymbol{Z}_I  & \text{otherwise}.
\end{cases}
\end{align}
Finally, patch features ${\boldsymbol{X}_i^p}$ and context features ${\bar{\boldsymbol{C}}_i}$ are fused via the Perceiver \cite{icml-2021-perceiver} module to generate vision latents ${\boldsymbol{X}_i^v} \in {\mathbb{R}^{m \times d}}$, which integrate visual semantics, view position, and clinical context into a unified latent space. This design allows the model to leverage clinical context when available, while gracefully degrading to image-only reasoning when context is missing.

\subsection{Multi-Level Supervision Paradigm}
To optimize clinically grounded representation learning, we propose a multi-level supervision paradigm with three objectives: (1) \textit{Global Alignment via Hybrid-Positive Contrastive Learning}, which unifies single- and multi-positive contrastive learning within a unified framework to achieve both intra- and inter-modal semantic alignment; (2) \textit{Disease-Aware Cross-Modal Representation Learning}, which aligns image and report within a shared disease label space, enhancing visual representations with disease-specific semantics; (3) \textit{Fine-Grained Attention via Soft Gaze Guidance}, which treats radiologist's gaze as probabilistic priors, explicitly linking salient image regions to corresponding textual cues and embedding diagnostic attention into the learned representation space.

\textbf{(1) Global Alignment via Hybrid-Positive Contrastive Learning.} Chest X-ray studies may include either a single radiograph or multiple views that share the same report. Conventional contrastive learning \cite{oord-cpc-infonce} assumes one positive per anchor, neglecting this clinically natural one-to-many correspondence. To better reflect the study structure, we propose a hybrid-positive contrastive learning method that unifies single- and multi-view cases within one framework. For each study, all associated images are paired with the same report, forming multiple positives in multi-view cases and naturally reducing to the single-positive setting otherwise. Formally, given a mini-batch of size $b$, we denote the global representations of vision latents and report embeddings as $\boldsymbol{X}^g_i,\boldsymbol{R}^g_j \in {\mathbb{R}^{d}}$. The image-to-report similarity distribution $\boldsymbol{q}^{I2R} \in {\mathbb{R} ^ {b \times b}}$ is given by:
\begin{align}
{\boldsymbol{q}}^{I2R}_{ij} = \frac{\exp(\text{sim}(\boldsymbol{X}^g_i, \boldsymbol{R}^g_j)/\tau_1)}{\sum_{k} \exp(\text{sim}(\boldsymbol{X}^g_i, \boldsymbol{R}^g_k)/\tau_1)},
\end{align}
where $\text{sim}(\cdot,\cdot)$ denotes cosine similarity and $\tau_1$ is a learnable temperature. Inspired by \cite{oord-cpc-infonce,NEURIPS2023_stablerep}, we define a categorical ground-truth distribution ${\boldsymbol{p}} \in {\mathbb{R}^{b \times b}}$ to encode study-level correspondences between image $x_i$ and report $y_j$. Specifically, ${\mathbbm{1}\{\text{study}(x_i)=\text{study}(y_j)\}}=1$ if they belong to the same study, and $0$ otherwise. To obtain a valid distribution, each row ${\boldsymbol{p}}_{i}$ is normalized by the number of positives:
\begin{align}
{\boldsymbol{p}}_{ij} = \frac{\mathbbm{1}\{\text{study}(x_i)=\text{study}(y_j)\}}{\sum_{k} \mathbbm{1}\{\text{study}(x_i)=\text{study}(y_k)\}}.
\end{align}
Finally, the hybrid-positive contrastive loss is the symmetric cross-entropy between ${\boldsymbol{p}}$ and ${\boldsymbol{q}}$:
\begin{align}
{{{\mathcal{L}}}_{con}} =  - \frac{1}{2b}\sum_{i}\sum_{j} {\left( {{\boldsymbol{p}}_{ij} \log {\boldsymbol{q}}_{ij}^{I2R} + {\boldsymbol{p}}_{ij}\log {\boldsymbol{q}}_{ij}^{R2I}} \right)}.
\end{align}
This formulation explicitly captures the one-to-many structure of clinical studies, enabling more consistent and semantically aligned vision-language representations.

\textbf{(2) Disease-Aware Cross-Modal Representation Learning.} To infuse diagnostic priors into visual representations, we present a disease-aware cross-modal representation learning framework that aligns image and report within a shared disease label space. Coupled with $\mathcal{L}_{con}$, this design encourages the vision encoder to capture disease-aware semantics, thereby enriching visual representations with diagnostic priors. Disease labels are derived from CheXbert \cite{Smit2020_chexbert}, which annotates 14 common thoracic observations. Specifically, \textit{No Finding} is binarized into two states, whereas the remaining 13 observations are represented using four states: blank, negative, uncertain, and positive. Following \cite{irvin-chexpert}, we treat blank as negative and uncertain as positive, converting all tasks into binary classification problems. To mitigate label imbalance across diseases, we employ a class-balanced focal loss \cite{2019-cvpr-class-balance-loss}:
\begin{align}
{\mathcal{L}}_{cls}^{M} = \frac{1-\beta}{1-{\beta}^{w_{\ell}}} \text{FL}(\text{logits}_M, \ell),
\end{align}
where $M \in \{V,T\}$ denotes the modality, and $\beta \in [0,1)$ is a hyperparameter, ${w_{\ell}}$ represents the number of positive samples in class $\ell$. $\text{FL}(\cdot, \cdot)$ denotes the focal loss \cite{2017-iccv-focal-loss} applied to the modality-specific logits. The final cross-modal classification objective is obtained by averaging the two modalities:
\begin{align}
\mathcal{L}_{cls} = 0.5 \times (\mathcal{L}_{cls}^V + \mathcal{L}_{cls}^T).
\end{align}

\textbf{(3) Fine-Grained Attention via Soft Gaze Guidance.} Learning fine-grained representation is crucial for accurate chest X-ray interpretation \cite{wang-mgca,cheng-prior,2025-cvpr-chexworld}. Prior methods \cite{wang-mgca,sei} rely on token-wise alignment but lack explicit clinical supervision. In contrast, radiologists' gaze provides direct cues to diagnostic focus. Nevertheless, gaze data is inherently noisy (e.g., head motion artifacts) and sparse, being available only for a limited number of cases, which poses challenges for direct supervision.

To address these issues, we propose a soft gaze guidance strategy (Fig.~\ref{fig:2-overview}(A)) that treats gaze as a probabilistic prior for transcript-patch alignment. We first compute transcript-to-patch similarity:
\begin{align}
\boldsymbol{S}^{t2p}_i =\text{sim} (\boldsymbol{T}^g_i,\boldsymbol{X}^p_i)/ \tau_2 \in \mathbb{R}^{n \times p},
\end{align}
where $\boldsymbol{T}^g_i \in \mathbb{R}^{n \times d}$ denotes the global transcript features aggregated from sentence- and paragraph-level embeddings. $n$ is the number of segments, $p$ the number of patches, and $\tau_2$ a learnable temperature. For supervision, raw gaze trajectories are first filtered to retain stable fixations \cite{ma2024-nips-eye-egma} and converted into heatmaps using a multivariate normal distribution. These heatmaps are then resized to the vision encoder's input resolution, pooled to patch granularity, and masked outside fixation regions. To reduce spurious signals from low-intensity areas, we retain only the top-$\rho$ fraction of non-zero elements in each heatmap, thereby sharpening supervision toward diagnostic focus. The resulting maps are normalized into probability distributions $\bar{\boldsymbol{G}}_i \in \mathbb{R}^{n \times p}$. Transcript-to-patch similarities are softmax-normalized to yield ${\bar{\boldsymbol{S}}}^{t2p}_i$. The soft gaze guidance loss is defined via a bidirectional Jensen-Shannon divergence (JSD):
\begin{align}
\label{eq-gaze}
\mathcal{L}_{G}^i = \lambda \cdot \text{JS}(\bar{\boldsymbol{G}}_i \parallel \bar{\boldsymbol{S}}^{t2p}_i) + (1-\lambda ) \cdot \text{JS}(\bar{\boldsymbol{G}}_i^T \parallel \bar{\boldsymbol{S}}^{p2t}_i),
\end{align}
where $\lambda$ balances transcript-to-patch (t2p) and patch-to-transcript (p2t) alignments. By modeling gaze as a probabilistic prior, our method assigns smoothly decaying weights from fixation points, in contrast to the uniform emphasis of EGMA \cite{ma2024-nips-eye-egma}. This formulation yields clinically grounded alignment, capturing diagnostic attention cues even under sparse supervision.

\textbf{Summary.} The pretraining objective comprises three components: ${\mathcal{L}}_{\text{pretrain}} = {\mathcal{L}}_{con} + {\mathcal{L}}_{cls }+{\mathcal{L}}_{G}.$ This encourages the model to learn semantically consistent and clinically grounded representations that generalize effectively across diverse downstream tasks.

\begin{figure}
\centering
\includegraphics[width=1\linewidth]{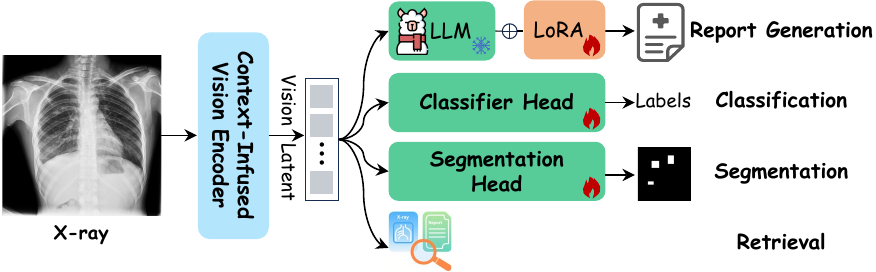}
\caption{Workflow of downstream tasks. All tasks are performed without requiring gaze heatmaps.}
\label{fig:downstream-tasks}
\end{figure}

\begin{table*}
	\centering
	\caption{Free-text report generation on MIMIC-CXR \cite{johnson-mimic-cxr-jpg} dataset. $\bigtriangleup$ denotes CoGaze's gain over the strongest baseline. EGMA \cite{ma2024-nips-eye-egma} is reproduced using its official code with DistilGPT2, while other results are from original papers ({\color{red}{\textbf{Best}}}, {\color{blue}{{{Second Best}}}}).}
	\label{tab:rg-result}
	\begin{tabular}{ccccccccccc} 
		\toprule
		\multirow{2}{*}{\textbf{Method}} & \multirow{2}{*}{\textbf{Venue}} & \multicolumn{6}{c}{\textbf{NLG Metrics} $\uparrow$} & \multicolumn{3}{c}{\textbf{CE Metrics} $\uparrow$} \\
		\cmidrule(lr){3-8}\cmidrule(lr){9-11}
		&  & \textbf{BLEU1} & \textbf{BLEU2} & \textbf{BLEU3}& \textbf{BLEU4} & \textbf{METEOR} & \textbf{R-L} & \textbf{P} & \textbf{R} & \textbf{$^{14}$Mi-F1 }\\ 
		\midrule
		KiUT \cite{huang-kiut} & CVPR'23 & 0.393 & 0.243 & 0.159 & 0.113 & 0.160 & 0.285 & 0.371 & 0.318 & 0.321 \\ 
		METransformer \cite{wang2023metransformer} & CVPR'23 & 0.386 & 0.250 & 0.169 & 0.124 & 0.152 & 0.291 & 0.364 & 0.309 & 0.311 \\ 
		MAN \cite{shen2024automatic_aaai} & AAAI'24 & 0.396 & 0.244 & 0.162 & 0.115 & 0.151 & 0.274 & 0.411 & 0.398 & 0.389 \\ 
		R2GenGPT \cite{wang-2023-r2gengpt} & Meta-Radio'23 & {0.411} & {0.267} & {0.186} & 0.134 & 0.160 & 0.297 & 0.392 & 0.387 & 0.389 \\ 
		Med-LLM \cite{liu2024in-context-acmmm} & MM'24 & - & - & - & 0.128 & 0.161 & 0.289 & 0.412 & 0.373 & 0.395 \\ 
		R2-LLM \cite{aaai-liu2024bootstrapping-llm} & AAAI'24 & 0.402 & 0.262 & 0.180 & 0.128 & 0.175 & 0.291 & 0.465 & \color{blue}0.482 & 0.473 \\ 
		SEI \cite{sei} & MICCAI'24 & 0.382 & 0.247 & 0.177 & {0.135} & 0.158 & 0.299 & 0.523 & 0.410 & 0.460 \\ 
		EGMA \cite{ma2024-nips-eye-egma} & NeurIPS'24 & 0.395 & 0.260 & 0.183 & 0.132 & 0.184 & 0.307 & 0.500 & 0.453 & 0.475 \\ 
		HERGen \cite{2024-eccv-hergen} & ECCV'24 & 0.395 & 0.248 & 0.169 & 0.122 & 0.156 & 0.285 & - & - & - \\ 
		MPO \cite{xiao2025radiology-mpo-aaai-2025} & AAAI'25 & \color{blue}{{0.416}} & {0.269} & {0.191} & {0.139} & 0.162 & {0.309} & 0.436 & 0.376 & 0.353 \\ 
		LLaVA-Med \cite{NEURIPS2023-llava-med} & NeurIPS'23 & 0.354 & - & - & 0.149 & - & 0.276 & - & - & 0.427 \\ 
		CheXagent \cite{chen2024-chexagent} & AAAI'24 & 0.169 & - & - & 0.047 & - & 0.215 & - & - & 0.393 \\ 
		MambaXray-VL-L \cite{Wang-2025-CVPR-CXPMRG-Bench} & CVPR'25 & \color{red}{\textbf{0.422}} & 0.268 & 0.184 & 0.133 & 0.167 & 0.289 & \color{red}{\textbf{0.561}} & 0.460 & {0.505} \\ 
		MLRG \cite{Liu-2025-CVPR-mlrg} & CVPR'25 & 0.411 & 0.277 & {0.204} & {0.158} & {0.176} & \color{blue}{{{0.320}}} & {0.549} & {0.468} & {0.505} \\ 
		\cmidrule(r){1-11}
		\textbf{CoGaze (DistilGPT2)} & Ours & 0.410 & \color{blue}{{0.290}} & \color{red}{{\textbf{0.220}}} & \color{red}{\textbf{0.175}} & \color{blue}{{{0.191}}} & \color{red}{\textbf{{0.326}}} & \color{blue}{0.555} & \color{red}{\textbf{0.498}} & \color{red}{\textbf{0.525}} \\  
		\textbf{CoGaze (Llama-3B)} & Ours & \color{red}{\textbf{0.422}} & \color{red}{\textbf{0.293}} & \color{blue}{{0.219}} & \color{blue}{{{0.171}}} & \color{red}{\textbf{{0.202}}} & 0.315 & {{{0.552}}} & 0.480 & \color{blue}{{{0.513}}} \\
		$\bigtriangleup (\%) \uparrow$ & - & +0.0 & +1.6 & +1.6 & +1.7 & +2.6 & +0.6 & -0.6 & +1.8 & +2.0 \\ 
		\bottomrule
	\end{tabular}
\end{table*}

\begin{table}
	\centering
	\caption{Structured report generation on SRRG-Findings \cite{delbrouck-etal-2025-srrg} dataset. ``RG'' denotes F1-RadGraph \cite{jain-radgraph}. $^\spadesuit$ and $^\heartsuit$ are DistilGPT2 and Llama-3B variants of CoGaze ({\color{red}{\textbf{Best}}}, {\color{blue}{{{Second Best}}}}).}
	\label{tab:srrg-result}
	\setlength{\tabcolsep}{1.4mm}
	\begin{tabular}{ccccccc} 
		\toprule
		\multirow{2}{*}{\textbf{Model}} & \multicolumn{3}{c}{\textbf{Base Metrics$\uparrow$}} & \multicolumn{3}{c}{\textbf{F1-SRR \cite{delbrouck-etal-2025-srrg} $\uparrow$}} \\ 
		\cmidrule(lr){2-4}\cmidrule(lr){5-7}
		& \textbf{BLEU} & \textbf{R-L} & \textbf{RG} & \textbf{P} & \textbf{R} & \textbf{F1} \\ 
		\midrule
		CheXagent \cite{chen2024-chexagent} & 1.80 & 19.65 & \color{blue}15.41 & \color{red}\textbf{ 77.12 } & 82.56 & 77.90 \\
		RaDialog \cite{midl-2025-radialog} & 1.28 & 17.53 & 13.82 & 69.48 & 70.12 & 69.76 \\
		\textbf{CoGaze$^\spadesuit$} & \color{blue}2.80 & \color{blue}20.23 & 14.23 & \color{blue}75.82 & \color{red}\textbf{85.61} & \color{red}\textbf{78.32} \\
		\textbf{CoGaze$^\heartsuit$} & \color{red}\textbf{3.00} & \color{red}\textbf{21.64} & \color{red}\textbf{15.53} & 74.83 & \color{blue}{85.56} & \color{blue}{78.07} \\
		\bottomrule
	\end{tabular}
\end{table}

\textbf{(4) Downstream Tasks.} The overall workflow is shown in Fig.~\ref{fig:downstream-tasks}. Following prior studies \cite{wang-mgca,zhou2023-iclr-MRM,zhou2024benchx,2025-cvpr-chexworld}, we attach task-specific heads to the context-infused vision encoder to support diverse objectives: a large language model for report generation, linear classifiers for disease classification, and a UNet decoder for segmentation. All tasks are initialized from the pretrained model (Fig.~\ref{fig:2-overview}) and are trained or evaluated without gaze supervision. Retrieval and zero-shot classification are conducted in a training-free manner, while the remaining tasks are trained with full supervision \cite{zhou2024benchx}. For report generation, we employ Llama-3.2-3B-Instruct \cite{liu2025spinquant-llama-3.2} as the language generator, yielding the CoGaze (Llama-3B) variant. We further adopt DistilGPT2 \cite{Sanh2019DistilBERTAD} as a lightweight alternative, resulting in the CoGaze (DistilGPT2) variant.

\section{Experiments}

\subsection{Experimental Settings}
\textbf{Pretraining Dataset.} We pretrain on the MIMIC-CXR training set \cite{johnson-mimic-cxr-jpg}, which comprises 240,422 chest X-ray images, including 1,711 cases with gaze annotations from EGD \cite{2020-eye-gaze-data} and REFLACX \cite{2022-reflacx}. Detailed statistics are provided in Appendix Tab.~A1.

\textbf{Downstream Datasets.} We evaluate six downstream tasks using seven public chest X-ray datasets. For \emph{report generation}, we use MIMIC-CXR \cite{johnson-mimic-cxr-jpg} for free-text report generation and SRRG-Findings \cite{delbrouck-etal-2025-srrg} for structured report generation. For \emph{classification}, we consider both multi-label and binary settings. NIH \cite{wang2017chestx-nih} includes 112,120 images annotated with 14 thoracic disease labels. Binary classification datasets include: SIIM \cite{siim-acr-pneumothorax-segmentation} for pneumothorax detection (12,047 cases), RSNA \cite{shih2019augmenting-rsna} for pneumonia identification (26,684 cases), and Shenzhen \cite{2023-shenzhencxr} for tuberculosis diagnosis (662 cases). For \emph{segmentation}, we adopt RSNA \cite{shih2019augmenting-rsna} and TBX11K \cite{tbx11k-cvpr-2020}, which provide pixel-level lesion masks for pneumonia and tuberculosis, respectively. For \emph{retrieval}, we construct MIMIC-5x200, following \cite{medclip_wang_2022,zhou2024benchx}, by sampling 200 cases each for five common diseases (\textit{Atelectasis}, \textit{Cardiomegaly}, \textit{Consolidation}, \textit{Edema}, and \textit{Pleural Effusion}) from the MIMIC-CXR test set. Data partitioning follows the official splits for MIMIC-CXR and SRRG-Findings, BenchX protocols \cite{zhou2024benchx} for SIIM/RSNA/NIH/TBX11K, and CheXWorld settings \cite{2025-cvpr-chexworld} for Shenzhen. Additional details are provided in Appendix Sec.~A.

\textbf{(2) Metrics.} For \emph{report generation}, we evaluate both natural language generation (NLG) and clinical efficacy (CE). CE metrics are computed from CheXbert's 14 observations \cite{Smit2020_chexbert} with micro-averaged Precision (P), Recall (R), and F1-score ($^{14}$Mi-F1). NLG quality is assessed using BLEU, METEOR, and ROUGE-L (R-L). For \emph{classification}, we report F1 and AUROC. For \emph{segmentation}, we employ the micro-averaged Dice score. For \emph{retrieval}, we report Precision@K (P@K) and Recall@K (R@K), considering reports with the same disease label as the query image as relevant.

\begin{figure}
\centering
\includegraphics[width=1\linewidth]{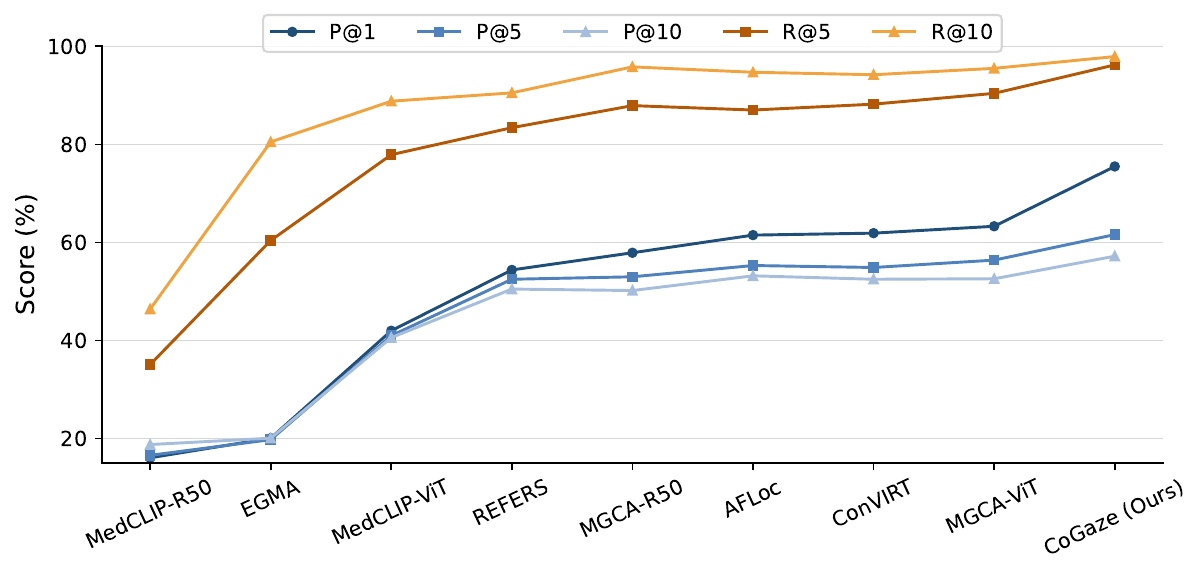}
\caption{Zero-shot image-text retrieval on MIMIC-5x200 dataset. P@K: precision at top-K; R@K: recall at top-K.}
\label{fig:retrieve}
\end{figure}

\begin{table*}
\centering
\caption{Classification results (\%) on NIH, SIIM, and RSNA datasets, reported as mean$\pm$std over three seeds ({\color{red}{\textbf{Best}}}, {\color{blue}{{{Second Best}}}}).}
\label{tab:cls-result}
\begin{tabular}{cccccccccc} 
\toprule
\multirow{2}{*}{\textbf{Model}} & \multicolumn{3}{c}{\textbf{NIH (AUROC$\uparrow$)}} & \multicolumn{3}{c}{\textbf{SIIM (F1$\uparrow$)}} & \multicolumn{3}{c}{\textbf{RSNA (F1$\uparrow$)}} \\ 
\cmidrule(lr){2-4}\cmidrule(lr){5-7}\cmidrule(lr){8-10}
& \textbf{1\%} & \textbf{10\%} & \textbf{100\%} & \textbf{1\%} & \textbf{10\%} & \textbf{100\%} & \textbf{1\%} & \textbf{10\%} & \textbf{100\%} \\ 
\midrule
MedCLIP-ViT \cite{medclip_wang_2022} & 76.1$\pm$0.3 & 81.4$\pm$0.25 & 84.5$\pm$0.17 & {68.6$\pm$0.8} & {71.5$\pm$1.1} & {75.7$\pm$0.2} & {63.5$\pm$0.5} & 65.3$\pm$1.0 & 66.2$\pm$0.8 \\
MedKLIP \cite{wu-medklip} & 75.2$\pm$0.1 & 80.3$\pm$0.08 & 83.9$\pm$0.08 & 61.4$\pm$0.3 & 64.4$\pm$2.1 & 72.7$\pm$1.4 & 60.4$\pm$0.6 & 61.9$\pm$1.4 & 66.0$\pm$0.6 \\
M-FLAG \cite{liu2023-miccai-m-flg} & 66.5$\pm$0.5 & 78.4$\pm$0.55 & 84.0$\pm$0.04 & 47.1$\pm$0.3 & 61.8$\pm$1.5 & 72.1$\pm$1.6 & 56.0$\pm$0.9 & 60.3$\pm$1.4 & 64.4$\pm$0.3 \\
MGCA-ViT \cite{wang-mgca} & 78.2$\pm$0.1 & 82.4$\pm$0.03 & 84.4$\pm$0.05 & 66.3$\pm$0.3 & 68.6$\pm$0.9 & 73.3$\pm$0.8 & 61.0$\pm$1.3 & 64.3$\pm$0.4 & 66.9$\pm$1.4 \\
MRM \cite{zhou2023-iclr-MRM} & \color{blue}{80.1$\pm$0.1} & \color{blue}{83.5$\pm$0.10} & \color{blue}{85.3$\pm$0.05} & 65.0$\pm$0.5 & 69.3$\pm$1.0 & 75.6$\pm$0.7 & 62.6$\pm$1.1 & {66.6$\pm$0.3} & 66.5$\pm$0.2 \\
REFERS \cite{zhou2022-NAI-REFERS} & 76.4$\pm$0.3 & 81.3$\pm$0.01 & 83.7$\pm$0.06 & 60.8$\pm$1.0 & 66.9$\pm$0.7 & 72.6$\pm$0.3 & 61.7$\pm$0.7 & 63.8$\pm$0.1 & {67.2$\pm$0.3} \\
EGMA \cite{ma2024-nips-eye-egma} & 66.2$\pm$1.2 & 73.9$\pm$1.29 & 81.8$\pm$0.42 & \color{blue}73.8$\pm$3.6 & 76.0$\pm$1.2 & \color{blue}97.1$\pm$0.3 & 79.9$\pm$0.5 & 82.5$\pm$0.3 & {84.4$\pm$0.2} \\
CheXWorld \cite{2025-cvpr-chexworld} & 60.5$\pm$1.4 & 68.8$\pm$1.82 & 79.0$\pm$0.60 & 53.1$\pm$2.3 & 75.4$\pm$2.2 & 95.9$\pm$0.4 & 80.3$\pm$1.0 & 81.4$\pm$0.3 & {84.1$\pm$0.1} \\
AFLoc \cite{2026-nbe-afloc} & 70.4$\pm$0.3 & 77.7$\pm$0.22 & 83.1$\pm$0.33 & 57.8$\pm$2.0 & \color{blue}78.0$\pm$0.5 & \color{red}\textbf{97.4$\pm$0.1} & \color{blue}81.5$\pm$0.8 & \color{blue}83.0$\pm$0.7 & \color{blue}{84.5$\pm$0.3} \\
\textbf{CoGaze (Ours)} & \color{red}{\textbf{80.7$\pm$0.2}} & \color{red}{\textbf{84.4$\pm$0.35}} & \color{red}{\textbf{86.1$\pm$0.12}} & \color{red}{\textbf{76.6$\pm$0.4}} & \color{red}{\textbf{78.2$\pm$0.3}} & \color{red}{\textbf{97.4$\pm$0.1}} & \color{red}{\textbf{83.3$\pm$0.4}} & \color{red}{\textbf{83.6$\pm$0.2}} & \color{red}{\textbf{84.8$\pm$0.4}} \\
\bottomrule
\end{tabular}
\end{table*}

\textbf{(3) Implementation Details.} We use CXR-BERT \cite{2022-eccv-cxr-bert} as the language encoder and Rad-DINO \cite{2024-rad-dino-nmi} as the vision encoder, with the number of latents set to $m=128$. Following CLIP \cite{radford-learning-clip}, the temperature parameters $\tau_1$ and $\tau_2$ are initialized as $\log(1/0.07)$. For transcript-to-patch alignment, we set $\lambda=0.8$ and retain the top 25\% ($\rho=0.25$) of non-zero heatmap elements to sharpen supervision. Additional details are outlined in Appendix Sec.~B.

\subsection{Downstream Tasks}
\textbf{Free-text Report Generation.} Tab.~\ref{tab:rg-result} compares CoGaze with 13 recent SOTA methods spanning five categories: (1) \textit{knowledge-graph approaches} (KiUT \cite{huang-kiut} and METransformer \cite{wang2023metransformer}); (2) \textit{LLM-based methods} (R2GenGPT \cite{wang-2023-r2gengpt}, Med-LLM \cite{liu2024in-context-acmmm}, and R2-LLM \cite{aaai-liu2024bootstrapping-llm}); (3) \textit{context- or temporal-aware models} (SEI \cite{sei} and HERGen \cite{2024-eccv-hergen}); (4) \textit{gaze-driven report generation} (EGMA \cite{ma2024-nips-eye-egma}); (5) \textit{reinforcement learning-based method} (MPO \cite{xiao2025radiology-mpo-aaai-2025}); (6) \textit{general and domain-specific vision-language models} (LLaVa-Med \cite{NEURIPS2023-llava-med}, MambaXray-VL-L \cite{Wang-2025-CVPR-CXPMRG-Bench}, and MLRG \cite{Liu-2025-CVPR-mlrg}). Across both NLG and CE metrics, CoGaze consistently outperforms general-purpose, domain-specific, and medical report generation models on MIMIC-CXR. The \textbf{Llama-3B variant} achieves higher scores on lower-order BLEUs, while the \textbf{DistilGPT2 variant} attains the best BLEU4, ROUGEL, and $^{14}$Mi-F1. Both variants yield the top $^{14}$Mi-F1 (0.525 and 0.513), indicating improvements in linguistic quality and clinical correctness.

\textbf{Structured Report Generation.} As presented in Tab.~\ref{tab:srrg-result}, CoGaze$^\heartsuit$ (Llama-3B) achieves the best overall results across all metrics, indicating superior clinical consistency and lexical similarity. CoGaze$^\spadesuit$ (DistilGPT2) attains the highest F1-SRR \cite{delbrouck-etal-2025-srrg} of 78.32\%, outperforming CheXagent \cite{chen2024-chexagent} by +0.42\%. Both variants outperform RaDialog \cite{midl-2025-radialog} across all metrics, confirming the effectiveness of CoGaze in generating structured and clinically faithful reports. 

\begin{figure}
\centering
\includegraphics[width=1\linewidth]{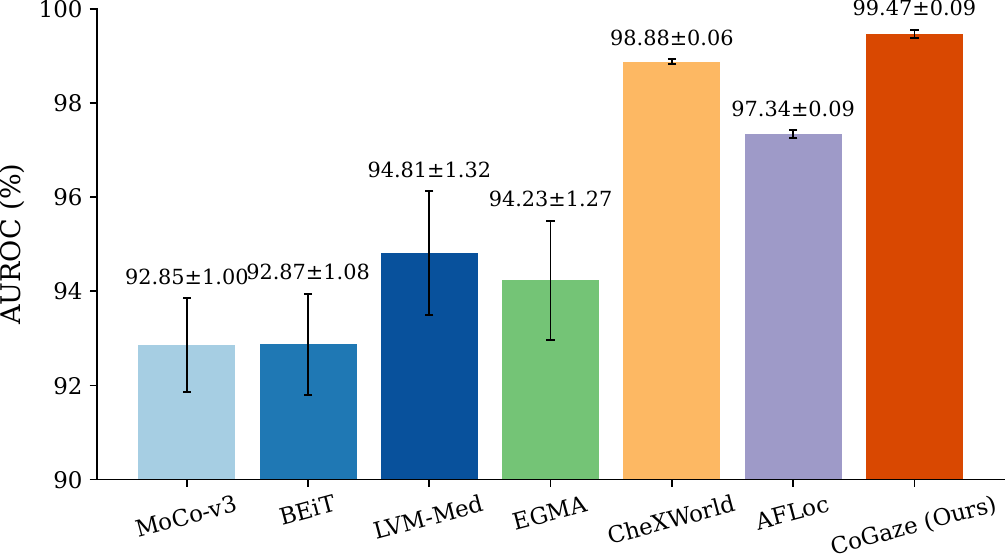}
\caption{Comparison of supervised classification performance in terms of AUROC on the Shenzhen \cite{2023-shenzhencxr} dataset.}
\label{afig-cls-shenzhen}
\end{figure}

\begin{table}
\centering
\caption{Zero-shot classification performance (\%) on RSNA \cite{shih2019augmenting-rsna} and Shenzhen \cite{2023-shenzhencxr} dataset. Since CheXWorld \cite{2025-cvpr-chexworld} lacks a text encoder, its results are omitted ({\color{red}{\textbf{Best}}}, {\color{blue}{{{Second Best}}}}).}
\label{tab:cls-zero-shot-result}
\begin{tabular}{ccccc} 
\toprule
\multirow{2}{*}{\textbf{ Method }} & \multicolumn{2}{c}{\textbf{ RSNA }} & \multicolumn{2}{c}{\textbf{ Shenzhen }} \\ 
\cmidrule(lr){2-3}\cmidrule(lr){4-5}
& \textbf{ F1$\uparrow$} & \textbf{ AUROC$\uparrow$} & \textbf{ F1$\uparrow$} & \textbf{ AUROC$\uparrow$} \\ 
\midrule
MedCLIP-ViT \cite{medclip_wang_2022} & 34.9 & 50.3 & 50.7 & 51.1 \\
MedKLIP \cite{wu-medklip} & 23.2 & 72.1 & 51.5 & 48.3 \\
M-FLAG \cite{liu2023-miccai-m-flg} & \color{red}\textbf{77.4} & 59.1 & 49.3 & 27.6 \\
MGCA-R50 \cite{wang-mgca} & 27.5 & 52.3 & 50.7 & 49.1 \\
MGCA-ViT \cite{wang-mgca} & 22.8 & 51.6 & 51.5 & 48.3 \\
MRM \cite{zhou2023-iclr-MRM} & 49.4 & 61.5 & 56.0 & \color{blue}71.5 \\
EGMA \cite{ma2024-nips-eye-egma} & 22.6 & 69.6 & 45.5 & 40.4 \\
AFLoc \cite{2026-nbe-afloc} & 67.7 & \color{blue}85.6 & \color{blue}76.1 & 58.0 \\
\textbf{ CoGaze (Ours) } & \color{blue}{ 77.0 } & \color{red}\textbf{ 86.2 } & \color{red}\textbf{ 81.3 } & \color{red}\textbf{ 94.7 } \\
\bottomrule
\end{tabular}
\end{table}

\textbf{Image-Text Retrieval.} As shown in Fig.~\ref{fig:retrieve}, CoGaze performs best on the MIMIC-5x200 dataset \cite{zhou2024benchx,johnson-mimic-cxr-jpg}. It attains a P@1 of 75.5\%, surpassing the strongest baseline, MGCA-ViT \cite{wang-mgca} (63.3\%), by +12.2 points, and outperforming ConVIRT \cite{pmlr-ConVIRT} and AFLoc \cite{2026-nbe-afloc} by +13.6 and +14.0 points, respectively. Gains remain consistent under less strict metrics, with improvements of +5.2 and +4.6 points on P@5 and P@10 over MGCA-ViT \cite{wang-mgca}. CoGaze further achieves 96.2\% and 97.9\% on R@5 and R@10, yielding gains of +5.8 and +2.4 points. These results indicate that CoGaze learns more discriminative and generalizable cross-modal representations, leading to consistently superior retrieval performance across all metrics.

\textbf{Supervised Classification.} Following \cite{zhou2024benchx,wang-mgca,zhou2023-iclr-MRM}, we evaluate classification performance under 1\%, 10\%, and 100\% labeled data settings. As shown in Tab.~\ref{tab:cls-result}, CoGaze consistently outperforms all baselines across NIH \cite{wang2017chestx-nih}, SIIM \cite{siim-acr-pneumothorax-segmentation}, and RSNA \cite{shih2019augmenting-rsna} datasets. It achieves the highest AUROC of 86.1\% on NIH and an F1 of 97.4\% on SIIM, demonstrating strong label efficiency and generalization. These results confirm that CoGaze effectively enhances representation quality under both limited- and full-supervision settings.

To ensure a fair comparison, we follow the data split protocol of CheXWorld~\cite{2025-cvpr-chexworld} for the Shenzhen dataset \cite{2023-shenzhencxr} and adopt their reported results for baseline methods, including MoCo-v3 \cite{chen2021-moco-v3}, BEiT \cite{bao2022beit}, LVM-Med \cite{nips-2023-lvm-med}, and CheXWorld \cite{2025-cvpr-chexworld}). EGMA \cite{ma2024-nips-eye-egma} and AFLoc \cite{2026-nbe-afloc} are reproduced using publicly available code or pretrained models. As shown in Fig.~\ref{afig-cls-shenzhen}, our CoGaze achieves the highest AUROC of 99.47\%, surpassing all competing methods. Compared with recent large-scale vision-language models such as CheXWorld~\cite{2025-cvpr-chexworld} and AFLoc \cite{2026-nbe-afloc}, CoGaze improves performance by +0.59\% and +2.13\%, respectively. These results highlight CoGaze's strong ability to capture disease-related visual cues.

\begin{table}
\centering
\caption{Segmentation performance (\%) on the RSNA \cite{shih2019augmenting-rsna} and TBX11K \cite{tbx11k-cvpr-2020} dataset ({\color{red}{\textbf{Best}}}, {\color{blue}{{{Second Best}}}}). }
\label{tab:seg-result}
\begin{tabular}{ccc} 
\toprule
\textbf{Method} & \textbf{RSNA (Dice$\uparrow$)} & \textbf{TBX11K (Dice$\uparrow$)} \\ 
\midrule
MedCLIP-R50 \cite{medclip_wang_2022} & 75.45$\pm$0.11 & 85.52$\pm$0.17 \\
MedCLIP-ViT \cite{medclip_wang_2022} & 73.29$\pm$1.41 & 85.62$\pm$0.07 \\
MedKLIP \cite{wu-medklip} & 74.68$\pm$0.42 & 87.06$\pm$0.31 \\
M-FLAG \cite{liu2023-miccai-m-flg} & 67.86$\pm$0.63 & 79.12$\pm$0.16 \\
MGCA-R50 \cite{wang-mgca} & 75.04$\pm$0.59 & 87.05$\pm$0.19 \\
MGCA-ViT \cite{wang-mgca} & 75.48$\pm$0.28 & 86.89$\pm$0.39 \\
MRM \cite{zhou2023-iclr-MRM} & 75.69$\pm$0.56 & 87.85$\pm$0.47 \\
REFERS \cite{zhou2022-NAI-REFERS} & 75.52$\pm$0.34 & 86.39$\pm$0.26 \\
EGMA \cite{ma2024-nips-eye-egma} & \color{blue}79.69$\pm$0.17 & \color{blue}95.86$\pm$0.12 \\
CheXWorld \cite{2025-cvpr-chexworld} & 75.52$\pm$0.34 & 86.39$\pm$0.26 \\
AFLoc \cite{2026-nbe-afloc} & 70.27$\pm$1.72 & 95.06$\pm$0.20 \\
\textbf{CoGaze (Ours)} & \color{red}\textbf{80.22$\pm$0.41} & \color{red}\textbf{96.56$\pm$0.11} \\
\bottomrule
\end{tabular}
\end{table}

\textbf{Zero-shot Classification.} As shown in Tab.~\ref{tab:cls-zero-shot-result}, CoGaze achieves superior performance on RSNA \cite{shih2019augmenting-rsna} and Shenzhen \cite{2023-shenzhencxr} datasets, with F1/AUROC of 77.0/86.2\% and 81.3/94.7\%, respectively. These results demonstrate effective transfer of visual-language knowledge to unseen domains and strong cross-dataset generalization.

\textbf{Segmentation.} We evaluate CoGaze on the RSNA \cite{shih2019augmenting-rsna} and TBX11K \cite{tbx11k-cvpr-2020} datasets for lesion segmentation. Tab.~\ref{tab:seg-result} shows that CoGaze attains the best Dice scores of 80.22\% and 96.27\%, outperforming the gaze-driven EGMA \cite{ma2024-nips-eye-egma}. These results suggest that CoGaze effectively strengthens spatial representation learning.

\begin{table}
\centering
\caption{Ablation study on supervision strategies, context integration, and gaze ratio for the MIMIC-CXR report generation task. ``BS'' and ``CC'' denotes BERTScore \cite{2020-iclr-bertscore} and clinical context, respectively. $^n$Mi-F1 and $^n$Ma-F1 refer to the micro- and macro-F1 scores computed by CheXbert, with $n$ observations. Higher is better for all metrics.}
\label{tab:ab-all}
\setlength{\tabcolsep}{0.38mm}
\begin{tabular}{lcccccc} 
\toprule
\textbf{Model} & \textbf{BLEU2} & \textbf{BS} & \textbf{\textsuperscript{14}Mi-F1} & \textbf{\textsuperscript{14}Ma-F1} & \textbf{\textsuperscript{5}Mi-F1} & \textbf{\textsuperscript{5}Ma-F1} \\ 
\midrule
\multicolumn{7}{c}{{\cellcolor[rgb]{0.802,1,1}}\textbf{\textit{Effect of context-infused vision encoder}}} \\ 
w/o CC & 0.210 & 0.535 & 0.499 & 0.358 & 0.540 & 0.467 \\ 
\cmidrule(r){1-7}
\multicolumn{7}{c}{{\cellcolor[rgb]{0.802,1,1}}\textbf{ \textit{Effect of multi-level supervision paradigm} }} \\ 
${\mathcal{L}_{con}}$ & 0.280 & 0.589 & 0.507 & 0.369 & 0.553 & 0.477 \\
${\mathcal{L}_{con}}$+${\mathcal{L}_{G}}$ & \color{blue}0.286 & \color{blue}0.595 & \color{blue}0.522 & 0.380 & \color{blue}0.568 & \color{red}\textbf{ 0.496 } \\ 
\cmidrule(lr){1-7}
\multicolumn{7}{c}{{\cellcolor[rgb]{0.802,1,1}}\textbf{ \textit{Effect of varying eye gaze ratio} }} \\ 
18 (1\%) Gaze & 0.282 & 0.591 & 0.489 & 0.348 & 0.527 & 0.459 \\
182 (10\%) Gaze & 0.283 & 0.593 & 0.507 & 0.371 & 0.549 & 0.482 \\
856 (50\%) Gaze & \color{blue}0.286 & \color{blue}0.595 & 0.517 & \color{blue}0.381 & 0.561 & 0.491 \\
\textbf{CoGaze(Ours)} & \color{red}\textbf{0.290} & \color{red}\textbf{ 0.596 } & \color{red}\textbf{ 0.525 } & \color{red}\textbf{ 0.388 } & \color{red}\textbf{ 0.571 } & \color{blue}0.495 \\
\bottomrule
\end{tabular}
\end{table}

\begin{table}
\caption{Ablation study on gaze-guidance losses, including mean squared error (MSE), intersection over union (IoU \cite{2024-accv-eye-cls}), and Jensen-Shannon divergence (JSD, i.e., CoGaze). ``ZSC'' denotes zero-shot classification on the RSNA dataset.}
\label{tab:gaze-constraints}
\setlength{\tabcolsep}{1.0mm}
\centering
\begin{tabular}{cccccccc} 
\toprule
\multirow{2}{*}{\textbf{Model}} & \textbf{ZSC} & \multicolumn{3}{c}{\textbf{Retrieval}} & \multicolumn{3}{c}{\textbf{Report Generation}} \\ 
\cmidrule(lr){2-2}\cmidrule(lr){3-5}\cmidrule(lr){6-8}
& \textbf{F1} & \textbf{P@1} & \textbf{P@5} & \textbf{P@10} & \textbf{BLEU2} & \textbf{\textsuperscript{14}Mi-F1} & \textbf{\textsuperscript{14}Ma-F1} \\ 
\midrule
MSE & 69.8 & 72.8 & 60.1 & 56.0 & 0.289 & 0.513 & 0.375 \\
IoU & 59.2 & 59.4 & 52.7 & 49.6 & 0.287 & 0.503 & 0.368 \\
\textbf{JSD} & \color{red}\textbf{77.0} & \color{red}\textbf{75.5} & \color{red}\textbf{61.6} & \color{red}\textbf{57.2} & \color{red}\textbf{0.290} & \color{red}\textbf{0.525} & \color{red}\textbf{0.388} \\
\bottomrule
\end{tabular}
\end{table}

\begin{figure}
\centering
\includegraphics[width=1\linewidth]{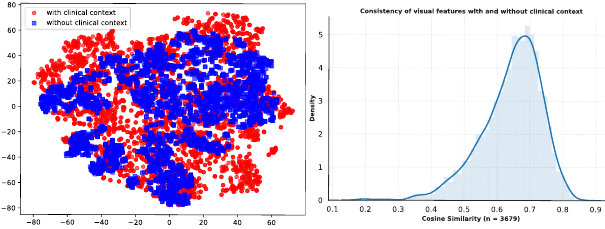}
\caption{t-SNE \cite{van2008visualizing} visualization (left) and cosine similarity distribution (right) of paired visual features extracted from the same image, with and without clinical context.}
\label{fig-clinical-context}
\end{figure}

\begin{table*}
\centering
\caption{Ablation study on hybrid-positive contrastive learning and gaze modeling strategies (\color{red}{\textbf{Best}}).}
\label{tab:ab-gaze-mask}
\setlength{\tabcolsep}{0.76mm}
\begin{tabular}{lccccccccccc} 
\toprule
\multirow{2}{*}{\textbf{Model}} & \multicolumn{4}{c}{\textbf{CLS (AUROC$\uparrow$)}} & \multicolumn{2}{c}{\textbf{SEG (Dice$\uparrow$)}} & \multicolumn{3}{c}{\textbf{Retrieval}} & \multicolumn{2}{c}{\textbf{Report Generation}} \\ 
\cmidrule(r){2-5}\cmidrule(r){6-7}\cmidrule(lr){8-10}\cmidrule(lr){11-12}
& \textbf{SIIM} & \textbf{RSNA} & \textbf{Shenzhen} & \textbf{NIH} & \textbf{RSNA} & \textbf{TBX11K} & \textbf{P@1$\uparrow$} & \textbf{P@5$\uparrow$} & \textbf{P@10$\uparrow$} & \textbf{BLEU2$\uparrow$} & \textbf{F1$\uparrow$} \\ 
\midrule
CoGaze w/ single-positive & 97.9 & 85.4 & 99.1 & 85.4 & 77.8 & 96.3 & 73.9 & 60.9 & 55.9 & 0.287 & 0.512 \\
CoGaze w/ gaze mask (EGMA \cite{ma2024-nips-eye-egma}) & 97.6 & 89.6 & 99.4 & 85.1 & 78.3 & 96.3 & 70.5 & 58.1 & 54.3 & 0.285 & 0.509 \\
\textbf{CoGaze (Ours)} & \color{red}\textbf{ 98.5 } & \color{red}\textbf{ 90.1 } & \color{red}\textbf{ 99.6 } & \color{red}\textbf{ 85.9 } & \color{red}\textbf{ 80.7 } & \color{red}\textbf{ 96.7 } & \color{red}\textbf{ 75.5 } & \color{red}\textbf{ 61.6 } & \color{red}\textbf{ 57.2 } & \color{red}\textbf{ 0.290 } & \color{red}\textbf{ 0.525 } \\
\bottomrule
\end{tabular}
\end{table*}

\begin{figure*}
	\centering
	\includegraphics[width=1\linewidth]{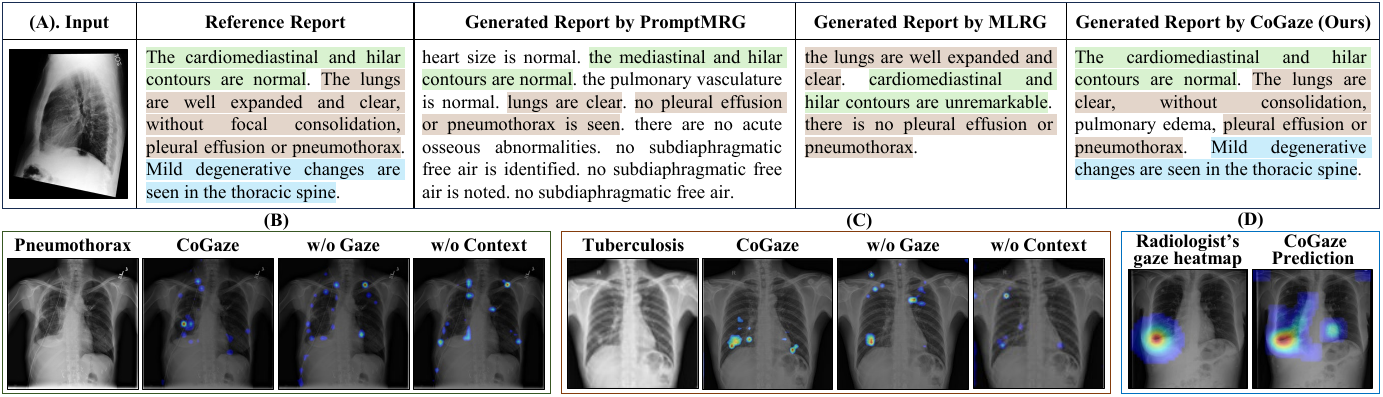}
	\caption{Qualitative results. (A) Free-text radiology reports generated from PromptMRG \cite{promtmrg-aaai-2024}, MLRG \cite{Liu-2025-CVPR-mlrg}, and CoGaze (DistilGPT2). (B-C) Attention maps of CoGaze, CoGaze w/o Gaze, and CoGaze w/o Context on pneumothorax and tuberculosis cases. (D) Comparison between radiologist' annotations and CoGaze-predicted heatmaps.}
	\label{fig:case-study}
\end{figure*}

\subsection{Ablation Study}
\textbf{Effect of Multi-Level Supervision Paradigm.} As shown in Tab.~ \ref{tab:ab-all}, starting from the hybrid-positive contrastive loss ${\mathcal{L}_{con}}$, incorporating gaze supervision (${\mathcal{L}_{con}}$+${\mathcal{L}_{G}}$) consistently improves performance across all metrics, with gains of + 0.6\% BLEU2, +1.5\% $^{14}$Mi-F1, and +1.1\% $^{5}$Mi-F1. This suggests that soft gaze guidance provides complementary fine-grained alignment beyond contrastive learning. Further introducing the classification loss ${\mathcal{L}_{cls}}$, the full model (CoGaze) achieves the best overall performance, reaching 0.290 BLEU2, 0.596 BERTScore, and 0.525/0.571 on $^{14}$Mi-F1 and $^{5}$Mi-F1, respectively. These results validate the effectiveness of the proposed multi-level supervision paradigm.

\textbf{Effect of Context-Infused Vision Encoder.} As presented in Tab.~ \ref{tab:ab-all}, removing clinical context from the vision encoder (i.e., w/o CC) leads to degraded report generation performance. This drop highlights the importance of contextual cues, suggesting that integrating clinical context enables the model to capture patient-specific semantics and learn more discriminative visual representations.

\textbf{Effect of Varying Gaze Ratio.} As illustrated in Tab.~\ref{tab:ab-all}, we vary the proportion of gaze-supervised samples from 1\% (18 samples) to 100\% (1,711 samples). Performance improves consistently across all metrics as the gaze ratio increases. For instance, $^{14}$Mi-F1 increases from 0.489$\rightarrow$0.507$\rightarrow$0.517$\rightarrow$0.525. Notably, even a small amount of gaze supervision (1,711 samples in total, corresponding to only $\sim$0.71\% of the pre-training data) provides effective fine-grained signals for vision-language alignment.

\textbf{Effect of Gaze-guidance Losses.} We replace Eq.~\ref{eq-gaze} with alternative objectives, including MSE and IoU (as in RET-GNN \cite{2024-accv-eye-cls}). As shown in Tab.~\ref{tab:gaze-constraints}, CoGaze (JSD) consistently performs best across zero-shot classification, image-text retrieval, and free-text report generation. These results suggest that JSD yields more informative and generalizable visual representations.

\textbf{Effect of Hybrid-Positive Contrastive Learning.} As shown in Tab.~\ref{tab:ab-gaze-mask}, CoGaze consistently outperforms its single-positive variant across all downstream tasks. By unifying single and multiple positives, it captures the one-to-many structure of clinical studies, thereby improving feature discrimination and generalization.

\textbf{Effect of Soft Gaze Guidance.} As reported in Tab.~\ref{tab:ab-gaze-mask}, modeling gaze as a probabilistic prior consistently outperforms the binary-mask variant used in EGMA \cite{ma2024-nips-eye-egma} across all tasks (i.e., CoGaze vs. CoGaze w/ gaze mask). This improvement arises because the soft gaze supervision assigns higher weights to diagnostically relevant regions, rather than treating all areas uniformly, providing smoother and more informative attention guidance.
\label{gaze-modeling-schems}

\textbf{Effect of Hyperparameters $\lambda$ and $\rho$.}  As shown in Appendix Fig.~\ref{fig-ab-hyperparameters}, $\lambda=0.8$ achieves an optimal balance between bidirectional alignment objectives, while $\rho=0.25$ enhances gaze supervision by emphasizing salient regions and suppressing low-intensity noise.

\subsection{Qualitative Analysis} 


To investigate the influence of clinical context on visual representations, we sample 3,679 images from the MIMIC-CXR test set, all of which include clinical context. We compare visual features extracted with and without context in terms of distribution and pairwise cosine similarity (Fig.~\ref{fig-clinical-context}). We observe that: (1) the two feature distributions largely overlap in the t-SNE space, indicating similar global structure; (2) for each sample, the cosine similarity between features extracted with and without context is predominantly above 0.65, suggesting high consistency. These results indicate that the learned visual representations are robust to missing clinical context and remain stable across conditions.

To further analyze CoGaze qualitatively, we visualize free-text report generation, attention maps, and gaze prediction in Fig.~\ref{fig:case-study}. We highlight three key observations. (1) Words in generated reports that match the reference are highlighted with consistent colors; greater color diversity reflects broader coverage of clinical findings. CoGaze (DistilGPT2) produces concise yet clinically faithful reports, accurately capturing both normal findings and subtle abnormalities (e.g., ``\textit{Mild degenerative changes are seen in the thoracic spine}''), whereas prior models \cite{promtmrg-aaai-2024,Liu-2025-CVPR-mlrg} often miss such fine-grained details. (2) CoGaze generates sharper and more lesion-focused attention maps for pneumothorax (from SIIM \cite{siim-acr-pneumothorax-segmentation}) and tuberculosis (from Shenzhen \cite{2023-shenzhencxr}) than its ablated variants, indicating improved spatial localization. (3) The predicted gaze heatmaps closely align with radiologists' gaze patterns, suggesting that CoGaze effectively captures human visual attention during pretraining. Additional examples are provided in Appendix Sec.~D.6.

\section{Conclusion}
In this work, we proposed CoGaze, a context- and gaze-guided vision-language model for chest X-ray. By jointly encoding view positions, clinical context, and radiologists' gaze cues, CoGaze effectively captures patient-specific context, integrates diagnostic priors, and attends to diagnostically salient regions, closely reflecting the radiological reasoning process. Extensive experiments demonstrated consistent improvements across report generation, disease classification, segmentation, and image-text retrieval tasks. Further work will investigate organ-aware \cite{2025-ORID} and spatiotemporal \cite{2025-DDaTR-longitudinal} modeling to further advance semantic understanding and localization precision.%


\begin{acks}
The work was jointly supported by the National Natural Science Foundations of China [grant number: 62272364]; the Provincial Key Research and Development Program of Shaanxi [grant number: 2024GH-ZDXM-47]; the Higher Education Science Research Planning Project of China Association of Higher Education [grant number: 24PG0101]; the Open Project of Hubei Provincial Key Laboratory of Multimedia Network Communication Engineering.
\end{acks}

\appendix
\renewcommand{\thetable}{A\arabic{table}}
\renewcommand{\thefigure}{A\arabic{figure}}
\renewcommand{\thesection}{\Alph{section}}

\begin{table}
	\centering
	\caption{Statistics of the MIMIC-CXR dataset used for pretraining. Gaze annotations are sourced from EGD \cite{2020-eye-gaze-data} and REFLACX \cite{2022-reflacx}. Since REFLACX \cite{2022-reflacx} provides up to five gaze recordings per image, we retain all of them to preserve radiologists' prior knowledge, resulting in a sample count mismatch between Appendix Tab.~\ref{tab-pretrain-data} and Tab.~\ref{tab-downstream-data}.}
	\label{tab-pretrain-data}
	\begin{tabular}{ccccc} 
		\toprule
		\textbf{Split} & \textbf{\#Image} & \textbf{\#Report} & \textbf{Context} & \textbf{Gaze} \\ 
		\midrule
		Train & 240,422 & 150,957 & 234,568 (97.57\%) & 1,711 (0.71\%) \\
		Val & 2,117 & 1,182 & 2,063 (97.45\%) & 10 (0.47\%) \\
		\bottomrule
	\end{tabular}
	
\end{table}

\begin{table*}
	\centering
	\caption{Data distribution of downstream tasks.}
	\label{tab-downstream-data}
	\begin{tabular}{cccccc} 
		\toprule
		\textbf{ Dataset } & \textbf{ Task } & \textbf{ Train } & \textbf{ Val } & \textbf{ Test } & \textbf{ Split } \\ 
		\midrule
		MIMIC-CXR \cite{johnson-mimic-cxr-jpg} & Free-text Report Generation & 240,197 & 2,113 & 3,852 & official split \\
		SRRG-Findings \cite{delbrouck-etal-2025-srrg} & Structured Report Generation & 181,874 & 976 & 1,459 & official split \\
		NIH \cite{wang2017chestx-nih} & 14-class Classification & 78,468 & 11,219 & 22,433 & BenchX \cite{zhou2024benchx} \\
		SIIM \cite{siim-acr-pneumothorax-segmentation} & Binary Classification & 9,303 & 1,372 & 1,372 & BenchX \cite{zhou2024benchx} \\
		Shenzhen \cite{2023-shenzhencxr} & Binary Classification & 463 & 65 & 134 & CheXWorld \cite{2025-cvpr-chexworld} \\
		RSNA \cite{shih2019augmenting-rsna} & Binary Classification \& Segmentation & 18,678 & 4,003 & 4,003 & BenchX \cite{zhou2024benchx} \\
		TBX11K \cite{tbx11k-cvpr-2020} & Segmentation & 5,879 & 1,260 & 1,260 & BenchX \cite{zhou2024benchx} \\
		\bottomrule
	\end{tabular}
\end{table*}

%
%
%
%

\section{Datasets}
\label{a-datasets}

We evaluate CoGaze on seven datasets spanning diverse medical vision-language tasks, including free-text and structured report generation, zero-shot and supervised disease classification, segmentation, and image-text retrieval. Detailed descriptions are provided below, and summary statistics are listed in Appendix Tab.~\ref{tab-downstream-data}.

\begin{itemize}
	\item \textbf{MIMIC-CXR}~\cite{johnson-mimic-cxr-jpg}: A large-scale, publicly available dataset of paired chest X-rays and free-text radiology reports collected at Beth Israel Deaconess Medical Center between 2011 and 2016. It comprises 377,110 images and 227,827 reports. We use the official training split for pretraining, with data distribution details presented in Appendix Tab.~\ref{tab-pretrain-data}. MIMIC-CXR also serves as the benchmark for the free-text report generation task. 
	
	\item \textbf{SRRG-Findings}~\cite{delbrouck-etal-2025-srrg}: A structured radiology report dataset derived from MIMIC-CXR~\cite{johnson-mimic-cxr-jpg} and CheXpert Plus~\cite{chexpert-plus-2024}, where free-text reports were converted into standardized structured formats using GPT-4. Each report is organized into predefined anatomical categories, including \textit{Lungs and Airways}, \textit{Pleura}, \textit{Cardiovascular}, \textit{Hila and Mediastinum}, \textit{Tubes, Catheters, and Support Devices}, \textit{Musculoskeletal and Chest Wall}, \textit{Abdominal}, and \textit{Other}. Observations are presented as bullet-point findings, explicitly covering both positive and negative cases. This dataset is used for the structured report generation. 
	
	\item \textbf{NIH}~\cite{wang2017chestx-nih}: A large-scale chest X-ray dataset released by the National Institutes of Health, containing 14 disease categories such as \textit{Atelectasis}, \textit{Cardiomegaly}, and \textit{Effusion}. It is used for multi-label classification.
	
	\item \textbf{SIIM}~\cite{siim-acr-pneumothorax-segmentation}: A publicly available Kaggle dataset, containing chest radiographs annotated for the presence of pneumothorax. It is used for binary classification.
	
	\item \textbf{Shenzhen}~\cite{2023-shenzhencxr}: A publicly available dataset developed by the U.S. National Library of Medicine in collaboration with the Third People's Hospital of Shenzhen City and the Guangdong Medical College in China. It consists of tuberculosis-labeled images and is used for binary and zero-shot classification.
	
	\item \textbf{RSNA}~\cite{shih2019augmenting-rsna}: A dataset released by the Radiological Society of North America, comprising frontal chest radiographs annotated for pneumonia. It supports binary and zero-shot classification, as well as segmentation tasks.
	
	\item \textbf{TBX11K}~\cite{tbx11k-cvpr-2020}: A chest X-ray dataset focusing on tuberculosis localization, providing bounding-box annotations of lesion regions. It is used for the segmentation task.
	
	\item \textbf{Eye Gaze Datasets}: The gaze annotations are sourced from EGD \cite{2020-eye-gaze-data} and REFLACX \cite{2022-reflacx}, both built upon the MIMIC-CXR \cite{johnson-mimic-cxr-jpg} database. Following \cite{ma2024-nips-eye-egma}, we retain only fixation-related gaze data to reduce noise and ensure reliability. Each sample consists of gaze coordinates paired with sentence- and paragraph-level audio transcripts. Detailed statistics are summarized in Appendix Tab.~\ref{tab-pretrain-data}.
	
	\item \textbf{Pretraining Dataset for Baselines.} MedCLIP \cite{medclip_wang_2022}, MedKLIP \cite{wu-medklip}, M-FLAG \cite{liu2023-miccai-m-flg}, MGCA \cite{wang-mgca}, MRM \cite{zhou2023-iclr-MRM}, and REFERS \cite{zhou2022-NAI-REFERS} are pretrained on the MIMIC-CXR training set, following BenchX \cite{zhou2024benchx}. EGMA \cite{ma2024-nips-eye-egma} restricts pretraining to the subset of MIMIC-CXR with eye-tracking annotations. CheXWorld \cite{2025-cvpr-chexworld} utilizes non-lateral radiographs from MIMIC-CXR (approximately 230K images). AFLoc \cite{2026-nbe-afloc} is pretrained on a mixture of MIMIC-CXR, Quilt-1M \cite{ikezogwo2023quilt-1m}, and an additional private set of 26,028 retinal fundus images.

\end{itemize}

\section{Implementation Details}
\label{a-implementation-details}
\subsection{Evaluation Metrics}
\textbf{Free-text Report Generation.} Natural language generation (NLG) metrics are implemented using the pycocoevalcap \cite{chen2015microsoft-coco-caption} library to assess the lexical similarity between generated and reference reports. BERTScore \cite{2020-iclr-bertscore} is used to measure semantic similarity via contextualized token matching based on BERT embeddings. Clinical efficacy (CE) metrics are computed with the f1chexbert \cite{Smit2020_chexbert} library to evaluate clinical correctness and disease consistency. We report $^n$Mi-F1 and $^n$Ma-F1, denoting the micro- and macro-F1 scores computed by CheXbert \cite{chexpert-plus-2024} over $n$ observations. Specifically, $n=14$ corresponds to the full set of 14 CheXbert-labeled observations, while $n=5$ restricts evaluation to \textit{Cardiomegaly}, \textit{Edema}, \textit{Consolidation}, \textit{Atelectasis}, and \textit{Pleural Effusion}.

\textbf{Structured Report Generation.} BLEU and ROUGEL measure the lexical similarity between generated and reference structured reports. F1-RadGraph (RG) \cite{jain-radgraph} evaluates clinical consistency by comparing extracted entities and relations, while F1-SRR \cite{delbrouck-etal-2025-srrg} quantifies alignment based on SRR-BERT's abnormality predictions across 55 disease categories. All metrics are computed using the StructEval library.

\begin{figure}
	\centering
	\includegraphics[width=1\linewidth]{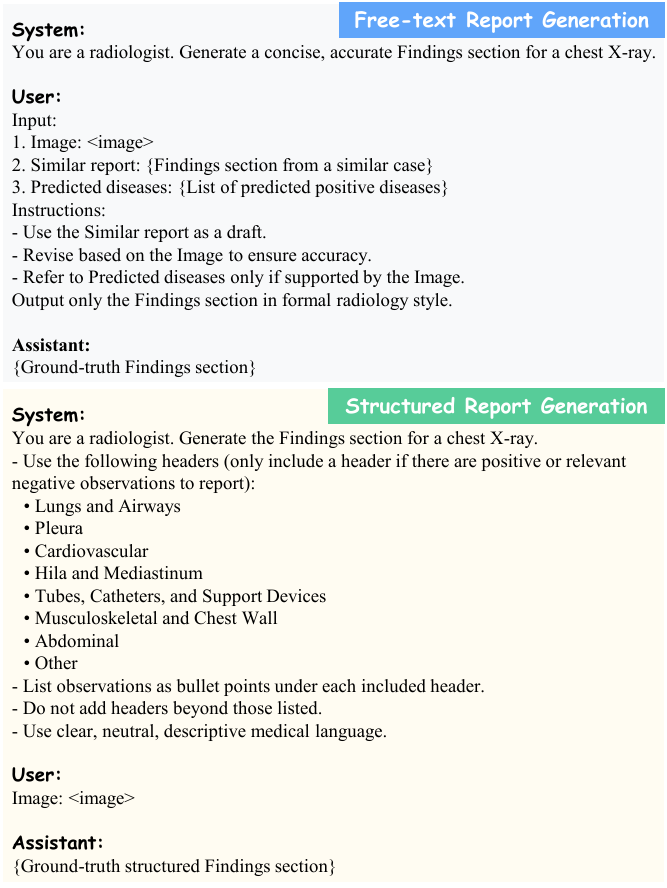}
	\caption{Prompts in CoGaze (Llama-3B) for free-text and structured report generation. For free-text report generation, a similar case is retrieved from the MIMIC-CXR training set based on vision latent similarity, and disease predictions are obtained from the vision-based classifier shown in Fig.~2.}
	\label{afig:rg-prompts}
\end{figure}

\begin{figure*}
	\centering
	\includegraphics[width=1\linewidth]{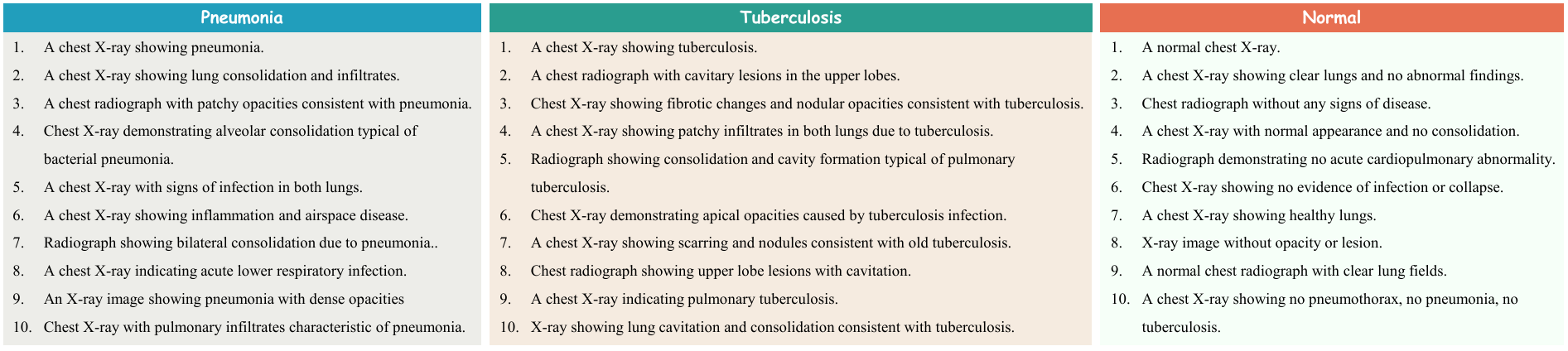}
	\caption{Category-specific prompts used for pneumonia, tuberculosis, and normal cases in zero-shot classification.}
	\label{afig:zero-shot-prompts}
\end{figure*}

\subsection{Baselines' Implementations}

\begin{itemize}
	\item EGMA \cite{ma2024-nips-eye-egma} provides only source code without released model weights; therefore, we reproduce its results on classification, segmentation, retrieval, and report generation tasks using the publicly available implementation.
	
	\item For CheXWorld \cite{2025-cvpr-chexworld} and AFLoc \cite{2026-nbe-afloc}, we reproduce results on classification, segmentation, and retrieval tasks using the provided source code and pretrained weights.
	
	\item For free-text report generation, baseline results are directly taken from the original publications. For structured report generation, we adopt results from SRR-BERT \cite{delbrouck-etal-2025-srrg}.
	
	\item To ensure fair comparison with prior medical vision-language pretraining methods, we adopt the classification, segmentation, and retrieval results of MedCLIP \cite{medclip_wang_2022}, MedKLIP \cite{wu-medklip}, M-FLAG \cite{liu2023-miccai-m-flg}, MGCA \cite{wang-mgca}, MRM \cite{zhou2023-iclr-MRM}, and REFERS \cite{zhou2022-NAI-REFERS} from BenchX \cite{zhou2024benchx}, a unified benchmark framework for chest X-ray vision-language pretraining.
	
	\item For the zero-shot classification task, we use model weights from BenchX (MedCLIP, MedKLIP, M-FLAG, MGCA, and MRM), the official release (AFLoc), and our reproduced implementation (EGMA), and evaluate all methods following the protocol described in Appendix Sec.~\ref{appendix-zero-shot-classification}.
	
	\item For the Shenzhen dataset, as BenchX does not provide an official data split, we follow the protocol of CheXWorld and adopt the same baselines, including MoCo-v3 \cite{chen2021-moco-v3}, BEiT \cite{bao2022beit}, LVM-Med \cite{nips-2023-lvm-med}, and CheXWorld \cite{2025-cvpr-chexworld}.
\end{itemize}

\subsection{CoGaze's Implementations}
We use the AdamW optimizer and a ReduceLROnPlateau learning rate scheduler for all experiments, conducted on a single NVIDIA RTX 5880 Ada GPU (48GB). The following sections describe implementation details for each downstream task, including pretraining, free-text and structured report generation, segmentation, and both supervised and zero-shot classification.

\subsubsection{Pretraining} We train our CoGaze for 10 epochs with a batch size of 80 and a learning rate of 5e-5. The model has approximately 225M parameters, of which 139M are trainable.

\subsubsection{Free-text and Structured Report Generation} For the CoGaze (DistilGPT2) variant, we use a learning rate of 5e-5 and train for up to 30 epochs. The model contains approximately 321M parameters, of which 235M are trainable. Decoding is performed with a beam size of 10. For the MIMIC-CXR dataset (free-text report generation), we use a batch size of 64 and a maximum output length of 100. For the SRRG-Findings dataset (structured report generation), we use a batch size of 48 and a maximum output length of 150.

For the CoGaze (LLaMA-3B) variant, we train for 10 epochs with a batch size of 6. The model has 3.4B parameters, with 6.9M trainable. The adapter is implemented as a single-layer MLP, and LoRA \cite{hu2022lora} is applied with a rank of 16, scaling factor $\alpha$ 16, and dropout rate of 0.1. The corresponding prompts are shown in Appendix Fig.~\ref{afig:rg-prompts}. We use a beam size of 3 and set the maximum output length to 100. The learning rate is 5e-5 on MIMIC-CXR and 1e-5 on SRRG-Findings.

\begin{table}
	\caption{Implementation details for segmentation (SEG) and supervised classification (CLS) tasks. ``LR'' denotes the learning rate. ``Patience'' indicates the number of consecutive epochs without improvement in validation loss before the learning rate scheduler decreases the learning rate.}
	\label{atab-cls-seg-details}
	\centering
	\setlength{\tabcolsep}{1.4mm}
	\begin{tabular}{cccccc} 
		\toprule
		\textbf{ Dataset } & \textbf{ Task } & \textbf{ Batch Size } & \textbf{LR} & \textbf{ Patience } & \textbf{ Epochs } \\ 
		\midrule
		RSNA & SEG & 16 & 1e-4 & 5 & 100 \\
		TBX11K & SEG & 16 & 5e-5 & 10 & 100 \\
		NIH & CLS & 16 & 1e-5 & 5 & 20 \\
		SIIM & CLS & 32 & 5e-6 & 2 & 50 \\
		RSNA & CLS & 16 & 1e-4 & 5 & 20 \\
		Shenzhen & CLS & 16 & 1e-5 & 5 & 50 \\
		\bottomrule
	\end{tabular}
\end{table}

\subsubsection{Segmentation and Supervised Classification} 
Appendix Tab.~\ref{atab-cls-seg-details} summarizes the batch size, learning rate, number of epochs, and the learning rate scheduler patience for each dataset.

\subsubsection{Zero-shot Classification} 
\label{appendix-zero-shot-classification}
For the RSNA \cite{shih2019augmenting-rsna} and Shenzhen \cite{2023-shenzhencxr} datasets, we construct category-specific prompts for three classes---pneumonia, tuberculosis, and normal. The full set of prompts is presented in Appendix Fig.~\ref{afig:zero-shot-prompts}. To enhance diversity and robustness, we design ten expert-reviewed prompts for each class, denoted as ${\mathcal{P}^c} = \{\mathcal{P}^c_k\}^{10}_{k=1}$. Following CLIP \cite{radford-learning-clip} and its extension \cite{zhou2022-ijcv-coop}, we employ a prompt ensemble strategy. Specifically, the textual embeddings of prompts within the same class are averaged to form a semantic prototype:
\begin{align}
	{\boldsymbol{\mathcal{{P}}}^c} = \frac{1}{10} \sum_{k=1}^{10} {\boldsymbol{\mathcal{{P}}}^c_k},
\end{align}
where ${\boldsymbol{\mathcal{{P}}}^c_k} \in {\mathbb{R}^d}$ denotes the global embeddings of the $k^{th}$ prompt ${\mathcal{{P}}}^c_k$ for class $c$, obtained from the language encoder. Visual features are extracted from the context-infused vision encoder, which is initialized with the pretrained model shown in Fig. 2. Zero-shot predictions are then computed by measuring cosine similarities between image features and each class prototype, assigning the label with the highest similarity score.

\section{Comparison of Existing Context- or Gaze-based Methods}
\label{sec:a-gaze-based-methods}
\subsection{Comparison of Existing Gaze-based Methods}
\textbf{Compared to EGMA}~\cite{ma2024-nips-eye-egma}, the most relevant gaze-based method, CoGaze consistently outperforms it across all evaluated tasks, including free-text report generation, image-text retrieval, classification, and segmentation. Specifically, for free-text report generation, CoGaze improves BLEU2 and CheXbertF1 by 3.3\% and 5.0\%, respectively. For image-text retrieval, it achieves substantial gains of 55.4\% in Precision@1 and 35.8\% in Recall@5. In supervised classification on the NIH dataset, AUROC improves by 4.3\%, while in zero-shot classification on the Shenzhen dataset, F1 increases by 35.8\%. For segmentation on the TBX11K dataset, CoGaze further improves Dice by 0.5\%.

\begin{table}
	\caption{Comparison of CXR-VLM-EyeGaze \cite{2024-miccai-eye-vlm} and CoGaze in terms of model characteristics and report generation performance. ``View Position'' indicates the supported imaging views, and ``Gaze'' denotes whether eye-tracking data is required during free-text report generation.}
	\label{atab-gaze-based-methods}
	\centering
	\setlength{\tabcolsep}{0.6mm}
	\begin{tabular}{ccccc} 
		\toprule
		\textbf{Model} & \textbf{\#Param} & \textbf{View Position} & \textbf{Gaze} & \textbf{ROUGE-L} \\ 
		\midrule
		CXR-VLM-EyeGaze & 7B & PA & \ding{51} & 0.298 \\
		CoGaze & 321M & PA/AP/Lateral & \ding{55} & 0.326 \\
		\bottomrule
	\end{tabular}
\end{table}

\begin{figure*}
	\centering
	\includegraphics[width=1\linewidth]{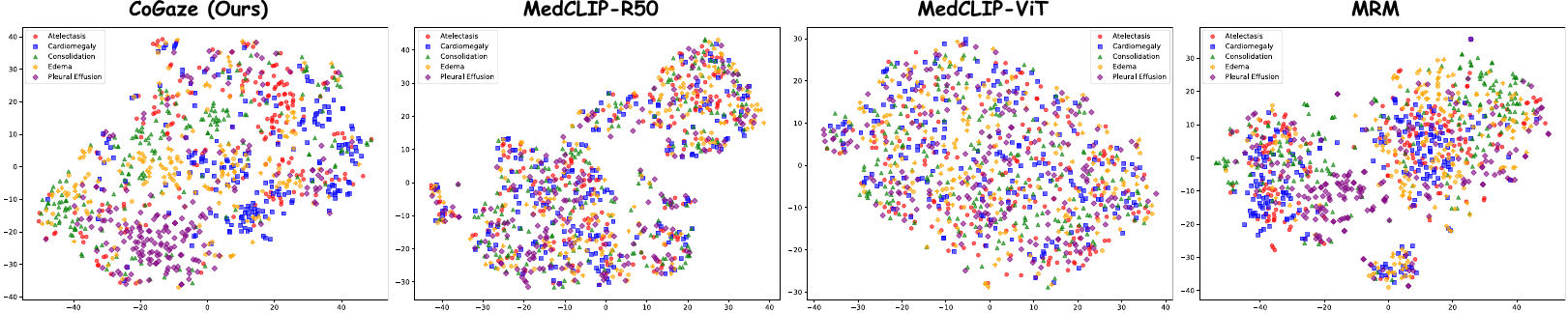}
	\caption{t-SNE \cite{van2008visualizing} visualization of the learned visual feature space on the MIMIC-5×200 dataset.}
	\label{afig:visual-feature-tsne}
\end{figure*}

\begin{table}
	\caption{Comparison between context-based method and CoGaze in supervised classification (CLS) and segmentation (SEG) performance.}
	\label{atab-context-based-methods}
	\centering
	\setlength{\tabcolsep}{0.6mm}
	\begin{tabular}{ccccc} 
		\toprule
		\multirow{2}{*}{\textbf{Model}} & \multicolumn{2}{c}{\textbf{CLS (AUROC$\uparrow$)}} & \multicolumn{2}{c}{\textbf{SEG (Dice$\uparrow$)}} \\ 
		\cmidrule(r){2-3}\cmidrule(lr){4-5}
		& \textbf{SIIM} & \textbf{Shenzhen} & \textbf{RSNA} & \textbf{TBX11K} \\ 
		\cmidrule(l){1-1}\cmidrule{2-5}
		PriorRG \cite{PriorRG} & 96.3$\pm$1.6 & 98.40$\pm$0.24 & 78.55$\pm$0.29 & 96.07$\pm$0.07 \\
		\textbf{ CoGaze } & \color{red}\textbf{ 97.4$\pm$0.1 } & \color{red}\textbf{ 99.47$\pm$0.09 } & \color{red}\textbf{ 80.22$\pm$0.41 } & \color{red}\textbf{ 96.56$\pm$0.11 } \\
		\bottomrule
	\end{tabular}
\end{table}

\textbf{Compared to CXR-VLM-EyeGaze}~\cite{2024-miccai-eye-vlm}, which does not release its source code and model weights, we conduct a comparison based on the reported model size and free-text report generation performance (Appendix Tab.~\ref{atab-gaze-based-methods}). CoGaze exhibits several key advantages. First, it adopts a significantly smaller model (321M vs. 7B parameters), resulting in improved computational efficiency and practicality. Second, CoGaze supports multiple view positions, including posteroanterior (PA), anteroposterior (AP), and lateral views, whereas CXR-VLM-Eyegaze is limited to PA images. Third, CoGaze does not require gaze signals during downstream tasks, in contrast to CXR-VLM-EyeGaze, which depends on gaze input at test time. Finally, CoGaze achieves a higher ROUGE-L score (0.326 vs. 0.298) in free-text report generation. Overall, these properties make CoGaze more suitable for real-world clinical scenarios, where diverse view positions are common, gaze annotations are often unavailable, and computational efficiency is critical.

\textbf{RET-GNN}~\cite{2024-accv-eye-cls} employs IoU as the gaze-guidance loss for chest X-ray classification; however, its source code and model weights are not publicly available. To enable a fair comparison between IoU and our Jensen-Shannon Divergence (JSD) objective, we replace Eq.~(8) with IoU within the CoGaze framework. As shown in Tab.~7, CoGaze with JSD consistently achieves the best performance across zero-shot classification, image-text retrieval, and free-text report generation. These results suggest that JSD leads to more informative and generalizable visual representations.

\subsection{Comparison of Existing Context-based Method}
We compare CoGaze with a representative context-based method, PriorRG~\cite{PriorRG}, on both supervised classification and segmentation tasks. As shown in Tab.~\ref{atab-context-based-methods}, CoGaze consistently outperforms PriorRG across all benchmarks. Specifically, for classification, CoGaze improves AUROC from 96.3 to 97.4 on SIIM and from 98.40 to 99.47 on the Shenzhen dataset. For segmentation, CoGaze achieves higher Dice scores on both RSNA (80.22 vs. 78.55) and TBX11K (96.56 vs. 96.07). These results indicate that CoGaze more effectively leverages contextual information, yielding consistent gains across both recognition and localization tasks.

\begin{figure}
	\centering
	\includegraphics[width=0.7\linewidth]{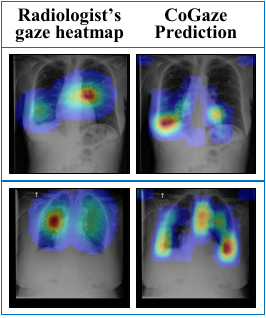}
	\caption{Comparison between radiologists' annotations and CoGaze-predicted heatmaps on the MIMIC-CXR dataset (Appendix Tab.~\ref{tab-pretrain-data}).}
	\label{afig:gaze-heatmap-consistency}
\end{figure}

\section{Additional Qualitative Analysis}
\label{sec:a-qualitative-analysis}

\begin{figure*}
	\centering
	\includegraphics[width=1\linewidth]{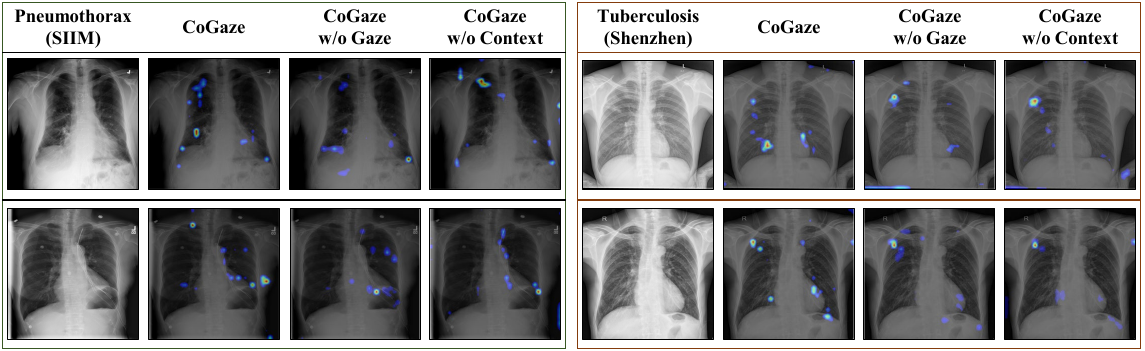}
	\caption{Attention visualizations of CoGaze, CoGaze w/o Gaze, and CoGaze w/o Context on pneumothorax (SIIM \cite{siim-acr-pneumothorax-segmentation}) and tuberculosis (Shenzhen \cite{2023-shenzhencxr}) cases.}
	\label{afig:attention-map}
\end{figure*}

\begin{figure*}
	\centering
	\includegraphics[width=1\linewidth]{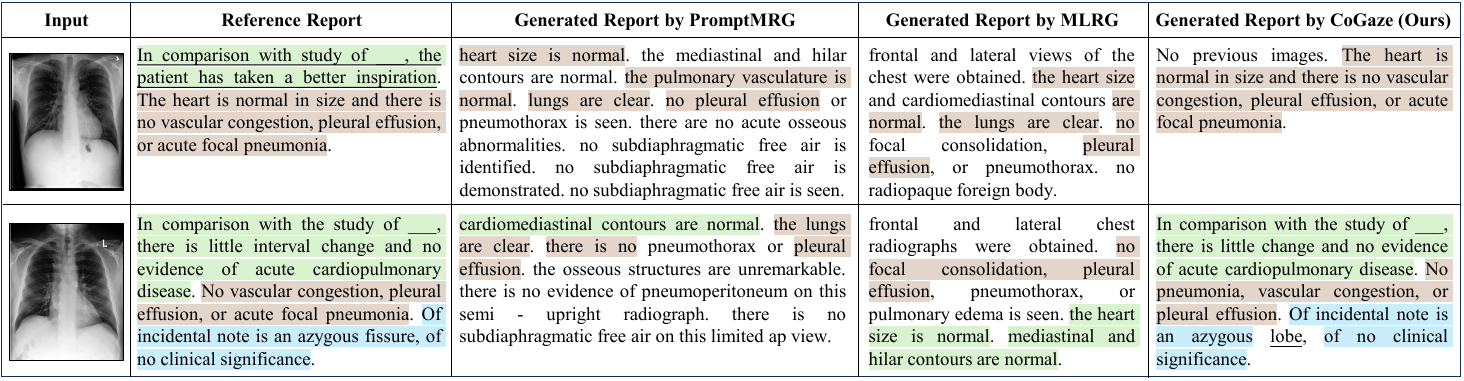}
	\caption{Examples of free-text radiology reports generated by PromptMRG \cite{promtmrg-aaai-2024}, MLRG \cite{Liu-2025-CVPR-mlrg}, and CoGaze (DistilGPT2). Words in the generated reports that match the reference are highlighted in the same color. Greater color diversity reflects broader coverage of clinical findings, while longer color spans suggest more detailed descriptions. Incorrect predictions are \underline{underlined}.}
	\label{afig:free-text-cases}
\end{figure*}

\subsection{Visual Feature Space Visualization on the MIMIC-5x200 Dataset} 
As shown in Appendix Fig.~\ref{afig:visual-feature-tsne}, we apply t-SNE \cite{van2008visualizing} to project the high-dimensional visual features into a 2D space. Compared to previous methods (i.e., MedCLIP-R50 \cite{medclip_wang_2022}, MedCLIP-ViT \cite{medclip_wang_2022}, and MRM \cite{zhou2023-iclr-MRM}), our CoGaze model produces clearer and more coherent clustering structures corresponding to the disease categories in the MIMIC-5×200 dataset \cite{zhou2024benchx}. This visualization suggests that CoGaze provides improved inter-class separability among the five disease categories.

\subsection{Comparison of Predicted and Radiologist Gaze Heatmaps} 
We evaluate the consistency between CoGaze-predicted heatmaps and radiologists' gaze heatmaps on the MIMIC-CXR validation set (Appendix Tab.~\ref{tab-pretrain-data}), using the model initialized with the pretrained weights described in Fig. 2. As illustrated in Appendix Fig.~\ref{afig:gaze-heatmap-consistency}, CoGaze consistently attends to regions aligned with radiologist gaze, indicating its ability to capture expert-like visual attention and highlight clinically meaningful areas.

\subsection{Attention Visualizations for Supervised Classification} 

Using models fine-tuned on 100\% of the training data, we visualize the attention maps via Grad-CAM \cite{2017-grad-cam} to interpret the model's decision process for pneumothorax (SIIM \cite{siim-acr-pneumothorax-segmentation}) and tuberculosis (Shenzhen \cite{2023-shenzhencxr}) cases (Appendix Fig.~\ref{afig:attention-map}). For pneumothorax, CoGaze primarily attends to the pleural margins and apical regions---areas typically associated with lung collapse and subpleural air accumulation. For tuberculosis, the model focuses on the apical and posterior segments of the upper lobes, as well as the superior segments of the lower lobes, consistent with the characteristic distribution of tuberculous lesions in chest radiographs. These findings suggest that CoGaze not only attains strong classification performance but also captures clinically meaningful attention patterns aligned with expert diagnostic reasoning.

\subsection{Examples of Free-text Report Generation} 
Appendix Fig.~\ref{afig:free-text-cases} presents qualitative comparisons on the MIMIC-CXR test set between PromptMRG \cite{promtmrg-aaai-2024}, MLRG \cite{Liu-2025-CVPR-mlrg}, and CoGaze (DistilGPT2). Compared to prior methods, CoGaze produces concise and clinically accurate reports that require minimal post-editing. For instance, in Case 1, only ``\textit{the patient has taken a better inspiration}'' needs to be added, while in Case 2, ``\textit{azygous lobe}'' can be corrected to ``\textit{azygous fissure}''. In contrast, existing methods produce longer reports that are less precise and often contain redundant or missing clinical details.

\begin{figure*}
	\centering
	\includegraphics[width=1\linewidth]{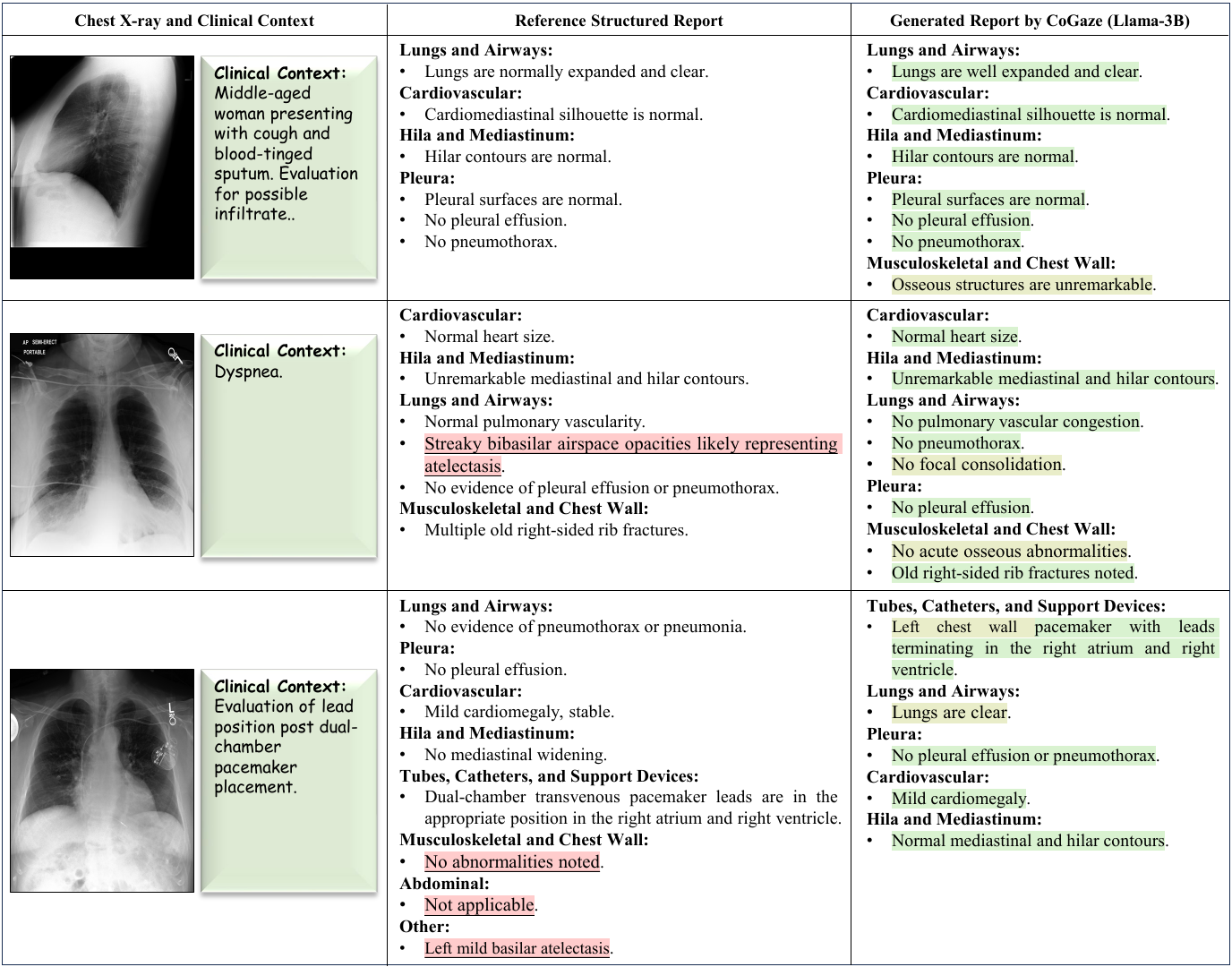}
	\caption{Examples of structured radiology reports generated by CoGaze (Llama-3B). Correctly generated words are highlighted in green, while acceptable words are highlighted in orange. Incorrect or missing findings are underlined in red.}
	\label{afig:structured-cases}
\end{figure*}

\begin{figure*}
	\centering
	\includegraphics[width=1\linewidth]{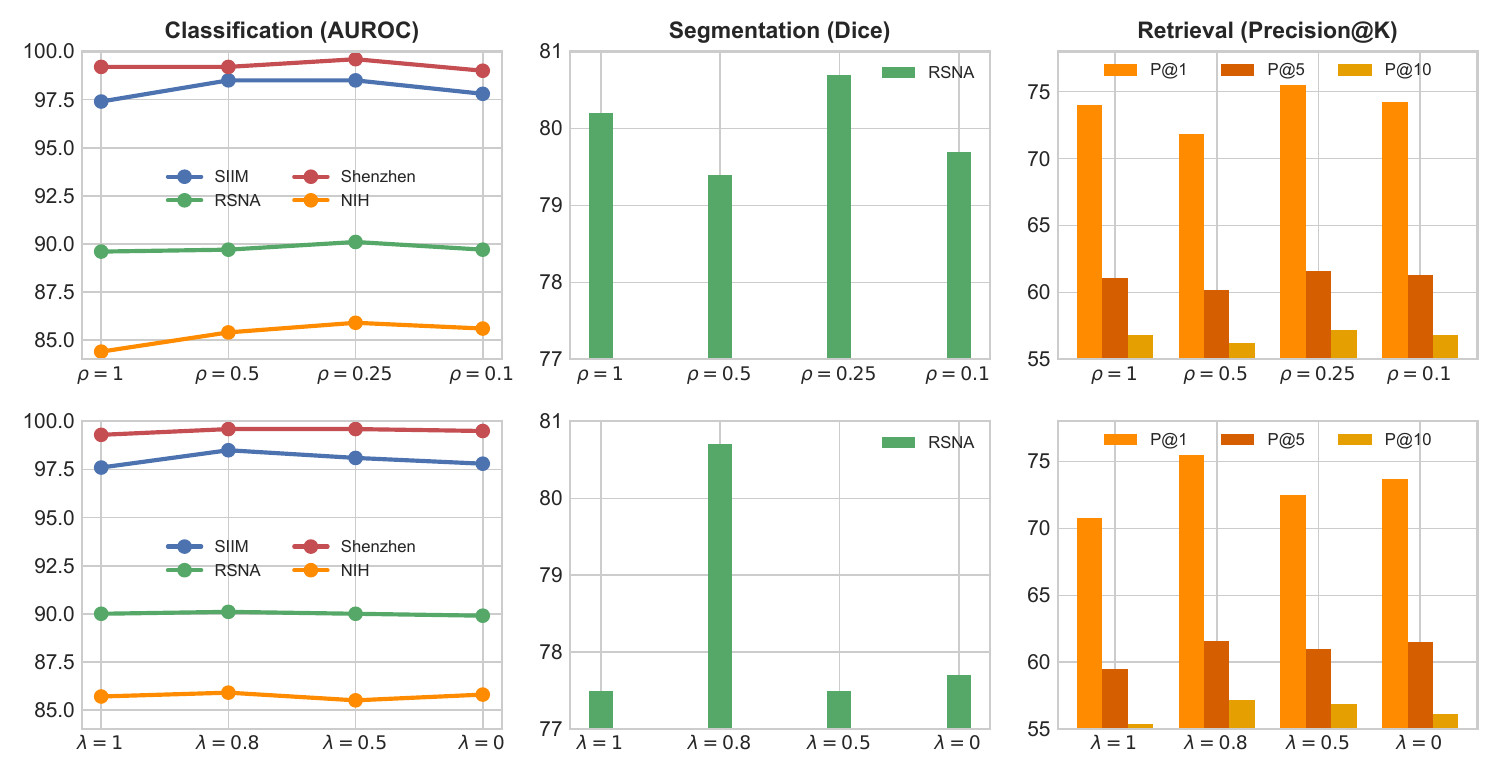}
	\caption{Ablation study on the impact of hyperparameters $\lambda$ and $\rho$ across classification (AUROC), segmentation (Dice), and retrieval (P@K) tasks on multiple datasets. The retrieval task is conducted on the MIMIC-5x200 \cite{johnson-mimic-cxr-jpg,zhou2024benchx} dataset. Our CoGaze achieves optimal performance at $\lambda=0.8$ and $\rho=0.25$.}
	\label{fig-ab-hyperparameters}
\end{figure*}

\subsection{Examples of Structured Report Generation} 
Appendix Fig.~\ref{afig:structured-cases} presents three examples from the SRRG-Findings \cite{delbrouck-etal-2025-srrg} test set. Our CoGaze (Llama-3B) accurately identifies primary findings with high factual accuracy and specificity. In particular, Case~1 requires almost no post-editing by radiologists, Case~2 correctly detects right-sided rib fractures (``\textit{Old right-sided rib fractures noted}''), and Case~3 precisely describes pacemaker placement and lead positions (``\textit{Left chest wall pacemaker with leads terminating in the right atrium and right ventricle}'').

\subsection{Failure Case Analysis for Report Generation}
\label{sec:fail-case}
In the free-text report generation task (Appendix Fig.~\ref{afig:free-text-cases}), CoGaze fails to generate the phrase ``\textit{the patient has taken a better inspiration}'' in Case 1. This limitation stems from the absence of temporal or longitudinal modeling \cite{2024-eccv-hergen,Liu-2025-CVPR-mlrg,zhou2025reviewlongitudinalradiologyreport}, which restricts the model's ability to capture changes across sequential studies. In Case~2, CoGaze mislabels the normal variant ``\textit{Azygous fissure}'' as ``\textit{Azygous lobe}''; this minor error remains clinically acceptable.

For the structured report generation (Appendix Fig.~\ref{afig:structured-cases}), CoGaze occasionally omits descriptions of normal findings, such as ``\textit{No abnormalities noted}'' and ``\textit{Not applicable}'' (Case~3), which are clinically negligible. It also fails to capture subtle abnormalities, including ``\textit{streaky bibasilar airspace opacities likely representing atelectasis}'' (Case 2) and ``\textit{Left mild basilar atelectasis}'' (Case~3), indicating challenges in distinguishing minor from more pronounced findings. This limitation likely arises from the absence of explicit priors for modeling severity distinctions. To address this, we are exploring attributed abnormality graphs \cite{yan_tmi_2023,tmi-2024-scene-graph-rrg} to better represent attribute-specific disease states.

\bibliographystyle{ACM-Reference-Format}
\bibliography{sample-base-abbrev}

@STRING{mar = "March"}

@STRING{jun = "June"}

@STRING{jul = "July"}

@STRING{aug = "Aug."}

@STRING{health = "ACM Transactions on Computing for Healthcare"}

@String{Computer = "{IEEE} Computer" }

@String{Springer = "Springer-Verlag" }

@article{lecun2015deep,
  title={Deep learning},
  author={LeCun, Yann and Bengio, Yoshua and Hinton, Geoffrey},
  journal={Nature},
  volume={521},
  number={7553},
  pages={436--444},
  year={2015},
  publisher={Nature Publishing Group UK London}
}

@ARTICLE{tpami-foundation-survey,
  author={Awais, Muhammad and Naseer, Muzammal and Khan, Salman and Anwer, Rao Muhammad and Cholakkal, Hisham and Shah, Mubarak and Yang, Ming-Hsuan and Khan, Fahad Shahbaz},
  journal={IEEE Transactions on Pattern Analysis and Machine Intelligence}, 
  title={Foundation Models Defining a New Era in Vision: A Survey and Outlook}, 
  year={2025},
  volume={47},
  number={4},
  pages={2245-2264},
  doi={10.1109/TPAMI.2024.3506283}
}

@INPROCEEDINGS{tbx11k-cvpr-2020,
  author={Liu, Yun and Wu, Yu-Huan and Ban, Yunfeng and Wang, Huifang and Cheng, Ming-Ming},
  booktitle={CVPR}, 
  title={Rethinking Computer-Aided Tuberculosis Diagnosis}, 
  year={2020},
  volume={},
  number={},
  pages={2643-2652},
  doi={10.1109/CVPR42600.2020.00272}
}

@inproceedings{liu2023-miccai-m-flg,
  title={M-flag: Medical vision-language pre-training with frozen language models and latent space geometry optimization},
  author={Liu, Che and Cheng, Sibo and Chen, Chen and Qiao, Mengyun and Zhang, Weitong and Shah, Anand and Bai, Wenjia and Arcucci, Rossella},
  booktitle={MICCAI},
  pages={637--647},
  year={2023},
  organization={Springer}
}

@article{ma2025fully-ark-Nature,
  title={A fully open AI foundation model applied to chest radiography},
  author={Ma, DongAo and Pang, Jiaxuan and Gotway, Michael B and Liang, Jianming},
  journal={Nature},
  pages={1--11},
  year={2025},
  publisher={Nature Publishing Group UK London}
}

@InProceedings{Liu-2025-CVPR-mlrg,
    author    = {Liu, Kang and Ma, Zhuoqi and Kang, Xiaolu and Li, Yunan and Xie, Kun and Jiao, Zhicheng and Miao, Qiguang},
    title     = {Enhanced Contrastive Learning with Multi-view Longitudinal Data for Chest X-ray Report Generation},
    booktitle = {CVPR},
    month     = {June},
    year      = {2025},
    pages     = {10348-10359}
}

@InProceedings{Wang-2025-CVPR-CXPMRG-Bench,
    author    = {Wang, Xiao and Wang, Fuling and Li, Yuehang and Ma, Qingchuan and Wang, Shiao and Jiang, Bo and Tang, Jin},
    title     = {CXPMRG-Bench: Pre-training and Benchmarking for X-ray Medical Report Generation on CheXpert Plus Dataset},
    booktitle = {CVPR},
    month     = {June},
    year      = {2025},
    pages     = {5123-5133}
}

@inproceedings{wang2017chestx-nih,
  title={Chestx-ray8: Hospital-scale chest x-ray database and benchmarks on weakly-supervised classification and localization of common thorax diseases},
  author={Wang, Xiaosong and Peng, Yifan and Lu, Le and Lu, Zhiyong and Bagheri, Mohammadhadi and Summers, Ronald M},
  booktitle={CVPR},
  pages={2097--2106},
  year={2017}
}

@article{2023-shenzhencxr,
  title={Automatic tuberculosis screening using chest radiographs},
  author={Jaeger, Stefan and Karargyris, Alexandros and Candemir, Sema and Folio, Les and Siegelman, Jenifer and Callaghan, Fiona and Xue, Zhiyun and Palaniappan, Kannappan and Singh, Rahul K and Antani, Sameer and others},
  journal={IEEE Transactions on Medical Imaging},
  volume={33},
  number={2},
  pages={233--245},
  year={2013},
  publisher={IEEE}
}

@misc{siim-acr-pneumothorax-segmentation,
    author = {Anna Zawacki and Carol Wu and George Shih and Julia Elliott and Mikhail Fomitchev and Mohannad Hussain and ParasLakhani and Phil Culliton and Shunxing Bao},
    title = {SIIM-ACR Pneumothorax Segmentation},
    year = {2019},
    howpublished = {\url{https://kaggle.com/competitions/siim-acr-pneumothorax-segmentation}},
    note = {Kaggle}
}

@article{shih2019augmenting-rsna,
  title={Augmenting the national institutes of health chest radiograph dataset with expert annotations of possible pneumonia},
  author={Shih, George and Wu, Carol C and Halabi, Safwan S and Kohli, Marc D and Prevedello, Luciano M and Cook, Tessa S and Sharma, Arjun and Amorosa, Judith K and Arteaga, Veronica and Galperin-Aizenberg, Maya and others},
  journal={Radiology: Artificial Intelligence},
  volume={1},
  number={1},
  pages={e180041},
  year={2019},
  publisher={Radiological Society of North America}
}

@article{zhou2022-NAI-REFERS,
  title={Generalized radiograph representation learning via cross-supervision between images and free-text radiology reports},
  author={Zhou, Hong-Yu and Chen, Xiaoyu and Zhang, Yinghao and Luo, Ruibang and Wang, Liansheng and Yu, Yizhou},
  journal={Nature Machine Intelligence},
  volume={4},
  number={1},
  pages={32--40},
  year={2022},
  publisher={Nature Publishing Group UK London}
}

@ARTICLE{2025-foundation-model-medicine,
  author={Khan, Wasif and Leem, Seowung and See, Kyle B. and Wong, Joshua K. and Zhang, Shaoting and Fang, Ruogu},
  journal={IEEE Reviews in Biomedical Engineering}, 
  title={A Comprehensive Survey of Foundation Models in Medicine}, 
  year={2025},
  volume={},
  number={},
  pages={1-22},
  doi={10.1109/RBME.2025.3531360}
}

@InProceedings{pmlr-ConVIRT,
  title = 	 {Contrastive Learning of Medical Visual Representations from Paired Images and Text},
  author =       {Zhang, Yuhao and Jiang, Hang and Miura, Yasuhide and Manning, Christopher D. and Langlotz, Curtis P.},
  booktitle = 	 {ML4H},
  pages = 	 {2--25},
  year = 	 {2022},
  volume = 	 {182},
  month = 	 {05--06 Aug},
  publisher =    {PMLR},
}

@inproceedings{zhou2023-iclr-MRM,
    title={Advancing Radiograph Representation Learning with Masked Record Modeling},
    author={Hong-Yu Zhou and Chenyu Lian and Liansheng Wang and Yizhou Yu},
    booktitle={ICLR},
    year={2023},
    url={https://openreview.net/forum?id=w-x7U26GM7j}
}

@inproceedings{liu2025spinquant-llama-3.2,
    title={SpinQuant: {LLM} Quantization with Learned Rotations},
    author={Zechun Liu and Changsheng Zhao and Igor Fedorov and Bilge Soran and Dhruv Choudhary and Raghuraman Krishnamoorthi and Vikas Chandra and Yuandong Tian and Tijmen Blankevoort},
    booktitle={ICLR},
    year={2025},
    url={https://openreview.net/forum?id=ogO6DGE6FZ}
}

@article{huang2024-NC-MaCo,
  title={Enhancing representation in radiography-reports foundation model: A granular alignment algorithm using masked contrastive learning},
  author={Huang, Weijian and Li, Cheng and Zhou, Hong-Yu and Yang, Hao and Liu, Jiarun and Liang, Yong and Zheng, Hairong and Zhang, Shaoting and Wang, Shanshan},
  journal={Nature Communications},
  volume={15},
  number={1},
  pages={7620},
  year={2024},
  publisher={Nature Publishing Group UK London},
  doi={10.1038/s41467-024-51749-0},
}

@inproceedings{cheng-prior,
   author = {Cheng, Pujin and Lin, Li and Lyu, Junyan and Huang, Yijin and Luo, Wenhan and Tang, Xiaoying},
   title = {PRIOR: Prototype Representation Joint Learning from Medical Images and Reports},
   booktitle = {ICCV},
   pages = {21361-21371},
   year = {2023},
   doi = {10.1109/ICCV51070.2023.01953},
}

@inproceedings{devlin-etal-2019-bert,
   author = {Devlin, Jacob and Chang, Ming-Wei and Lee, Kenton and Toutanova, Kristina},
   title = {BERT: Pre-training of Deep Bidirectional Transformers for Language Understanding},
   booktitle = {NAACL},
   volume = {1},
   pages = {4171-4186},
   DOI = {10.18653/v1/N19-1423},
   year = {2019},
}

@inproceedings{NEURIPS2023-llava-med,
     author = {Li, Chunyuan and Wong, Cliff and Zhang, Sheng and Usuyama, Naoto and Liu, Haotian and Yang, Jianwei and Naumann, Tristan and Poon, Hoifung and Gao, Jianfeng},
     booktitle = {NeurIPS},
     pages = {28541--28564},
     publisher = {Curran Associates, Inc.},
     title = {LLaVA-Med: Training a Large Language-and-Vision Assistant for Biomedicine in One Day},
     volume = {36},
     year = {2023}
}

@article{zambrano2025-llava-rad,
  title={A clinically accessible small multimodal radiology model and evaluation metric for chest X-ray findings},
  author={Zambrano Chaves, Juan Manuel and Huang, Shih-Cheng and Xu, Yanbo and Xu, Hanwen and Usuyama, Naoto and Zhang, Sheng and Wang, Fei and Xie, Yujia and Khademi, Mahmoud and Yang, Ziyi and others},
  journal={Nature Communications},
  volume={16},
  number={1},
  pages={3108},
  year={2025},
  publisher={Nature Publishing Group UK London}
}

@inproceedings{xiao2025radiology-mpo-aaai-2025,
  title={Radiology report generation via multi-objective preference optimization},
  author={Xiao, Ting and Shi, Lei and Liu, Peng and Wang, Zhe and Bai, Chenjia},
  booktitle={AAAI},
  volume={39},
  number={8},
  pages={8664--8672},
  year={2025}
}

@ARTICLE{2025-tmi-foundation-model-bbox,
  author={Wang, Fuying and Yu, Lequan},
  journal={IEEE Transactions on Medical Imaging}, 
  title={Scaling Chest X-ray Foundation Models from Mixed Supervisions for Dense Prediction}, 
  year={2025},
  volume={},
  number={},
  pages={1-1},
  doi={10.1109/TMI.2025.3589928}}

@misc{2024-eva-x-foundation-model,
      title={EVA-X: A Foundation Model for General Chest X-ray Analysis with Self-supervised Learning}, 
      author={Jingfeng Yao and Xinggang Wang and Yuehao Song and Huangxuan Zhao and Jun Ma and Yajie Chen and Wenyu Liu and Bo Wang},
      year={2024},
      eprint={2405.05237},
      archivePrefix={arXiv},
      primaryClass={cs.CV},
}

@INPROCEEDINGS{2025-wacv-foundation-x,
  author={Islam, Nahid Ul and Ma, DongAo and Pang, Jiaxuan and Velan, Shivasakthi Senthil and Gotway, Michael and Liang, Jianming},
  booktitle={WACV}, 
  title={Foundation X: Integrating Classification, Localization, and Segmentation Through Lock-Release Pretraining Strategy for Chest X-Ray Analysis}, 
  year={2025},
  volume={},
  number={},
  pages={3647-3656},
  doi={10.1109/WACV61041.2025.00359}}

@inproceedings{2025-cvpr-chexworld,
  title={CheXWorld: Exploring Image World Modeling for Radiograph Representation Learning},
  author={Yue, Yang and Wang, Yulin and Tao, Chenxin and Liu, Pan and Song, Shiji and Huang, Gao},
  booktitle={CVPR},
  pages={20778--20788},
  year={2025}
}

@article{2024-rad-dino-nmi,
  title={Exploring scalable medical image encoders beyond text supervision},
  author={Perez-Garcia, Fernando and Sharma, Harshita and Bond-Taylor, Sam and Bouzid, Kenza and Salvatelli, Valentina and Ilse, Maximilian and Bannur, Shruthi and Castro, Daniel C and Schwaighofer, Anton and Lungren, Matthew P and others},
  journal={Nature Machine Intelligence},
  volume={7},
  number={1},
  pages={119--130},
  year={2025},
  publisher={Nature Publishing Group UK London}
}

@inproceedings{2022-eccv-cxr-bert,
  title={Making the most of text semantics to improve biomedical vision--language processing},
  author={Boecking, Benedikt and Usuyama, Naoto and Bannur, Shruthi and Castro, Daniel C and Schwaighofer, Anton and Hyland, Stephanie and Wetscherek, Maria and Naumann, Tristan and Nori, Aditya and Alvarez-Valle, Javier and others},
  booktitle={ECCV},
  pages={1--21},
  year={2022},
  organization={Springer}
}

@article{wang-2023-r2gengpt,
  title={R2gengpt: Radiology report generation with frozen llms},
  author={Wang, Zhanyu and Liu, Lingqiao and Wang, Lei and Zhou, Luping},
  journal={Meta-Radiology},
  volume={1},
  number={3},
  pages={100033},
  year={2023},
  publisher={Elsevier}
}

@ARTICLE{tmi-2024-scene-graph-rrg,
  author={Zhang, Ke and Yang, Yan and Yu, Jun and Fan, Jianping and Jiang, Hanliang and Huang, Qingming and Han, Weidong},
  journal={IEEE Transactions on Medical Imaging}, 
  title={Attribute Prototype-guided Iterative Scene Graph for Explainable Radiology Report Generation}, 
  year={2024},
  volume={},
  number={},
  pages={1-1},
  doi={10.1109/TMI.2024.3424505}
}

@ARTICLE{yan_tmi_2023,
    author={Yan, Sixing and Cheung, William K. and Chiu, Keith and Tong, Terence M. and Cheung, Ka Chun and See, Simon},
    journal={IEEE Transactions on Medical Imaging}, 
    title={Attributed Abnormality Graph Embedding for Clinically Accurate X-Ray Report Generation}, 
    year={2023},
    volume={42},
    number={8},
    pages={2211-2222},
    doi={10.1109/TMI.2023.3245608}
}

@inproceedings{medclip_wang_2022,
    author       = {Zifeng Wang and
                  Zhenbang Wu and
                  Dinesh Agarwal and
                  Jimeng Sun},
    title        = {MedCLIP: Contrastive Learning from Unpaired Medical Images and Text},
    booktitle    = {EMNLP},
    pages        = {3876-3887},
    year         = {2022},
    doi          = {10.18653/V1/2022.EMNLP-MAIN.256},
}

@inproceedings{promtmrg-aaai-2024,
    title = {PromptMRG: Diagnosis-Driven Prompts for Medical Report Generation},
    volume = {38},
    ISSN = {2159-5399},
    DOI = {10.1609/aaai.v38i3.28038},
    number = {3},
    booktitle = {AAAI},
    author = {Jin,  Haibo and Che,  Haoxuan and Lin,  Yi and Chen,  Hao},
    year = {2024},
    month = {Mar},
    pages = {2607–2615}
}

@inproceedings{aaai-liu2024bootstrapping-llm,
    title={Bootstrapping Large Language Models for Radiology Report Generation},
    author={Liu, Chang and Tian, Yuanhe and Chen, Weidong and Song, Yan and Zhang, Yongdong},
    booktitle={AAAI},
    volume={38},
    number={17},
    pages={18635-18643},
    doi={10.1609/aaai.v38i17.29826},
    year={2024}
}

@article{NEURIPS2023_stablerep,
  title={Stablerep: Synthetic images from text-to-image models make strong visual representation learners},
  author={Tian, Yonglong and Fan, Lijie and Isola, Phillip and Chang, Huiwen and Krishnan, Dilip},
  journal={NeurIPS},
  volume={36},
  year={2024}
}

@misc{2509-generative-foundation-model,
      title={A Generative Foundation Model for Chest Radiography}, 
      author={Yuanfeng Ji and Dan Lin and Xiyue Wang and Lu Zhang and Wenhui Zhou and Chongjian Ge and Ruihang Chu and Xiaoli Yang and Junhan Zhao and Junsong Chen and Xiangde Luo and Sen Yang and Jin Fang and Ping Luo and Ruijiang Li},
      year={2025},
      eprint={2509.03903},
      archivePrefix={arXiv},
      primaryClass={cs.CV},
}

@inproceedings{chen2024-chexagent,
    title={CheXagent: Towards a Foundation Model for Chest X-Ray Interpretation},
    author={Zhihong Chen and Maya Varma and Jean-Benoit Delbrouck and Magdalini Paschali and Louis Blankemeier and Dave Van Veen and Jeya Maria Jose Valanarasu and Alaa Youssef and Joseph Paul Cohen and Eduardo Pontes Reis and Emily Tsai and Andrew Johnston and Cameron Olsen and Tanishq Mathew Abraham and Sergios Gatidis and Akshay S Chaudhari and Curtis Langlotz},
    booktitle={AAAI 2024 Spring Symposium on Clinical Foundation Models},
    year={2024},
    url={https://openreview.net/forum?id=P3LOmrZWGR}
}

@inproceedings{Smit2020_chexbert,
    title = {Combining Automatic Labelers and Expert Annotations for Accurate Radiology Report Labeling Using BERT},
    DOI = {10.18653/v1/2020.emnlp-main.117},
    booktitle = {EMNLP},
    author = {Smit,  Akshay and Jain,  Saahil and Rajpurkar,  Pranav and Pareek,  Anuj and Ng,  Andrew and Lungren,  Matthew},
    year = {2020}
}

@inproceedings{shen2024automatic_aaai,
  title={Automatic Radiology Reports Generation via Memory Alignment Network},
  author={Shen, Hongyu and Pei, Mingtao and Liu, Juncai and Tian, Zhaoxing},
  booktitle={AAAI},
  volume={38},
  number={5},
  pages={4776-4783},
  year={2024}
}

@InProceedings{sei,
      author={Liu, Kang and Ma, Zhuoqi and Kang, Xiaolu and Zhong, Zhusi and Jiao, Zhicheng and Baird, Grayson and Bai, Harrison and Miao, Qiguang},
      title={Structural Entities Extraction and Patient Indications Incorporation for Chest X-Ray Report Generation},
      booktitle={MICCAI},
      year={2024},
      publisher={Springer Nature Switzerland},
      address={Cham},
      pages={433--443},
      isbn={978-3-031-72384-1},
      doi={10.1007/978-3-031-72384-1_41}
}

@misc{chen2015microsoft-coco-caption,
      title={Microsoft COCO Captions: Data Collection and Evaluation Server}, 
      author={Xinlei Chen and Hao Fang and Tsung-Yi Lin and Ramakrishna Vedantam and Saurabh Gupta and Piotr Dollar and C. Lawrence Zitnick},
      year={2015},
      eprint={1504.00325},
      archivePrefix={arXiv},
      primaryClass={cs.CV},
}

@inproceedings{zhou2024benchx,
  title={Benchx: A unified benchmark framework for medical vision-language pretraining on chest x-rays},
  author={Zhou, Yang and Faith, Tan and Xu, Yanyu and Leng, Sicong and Xu, Xinxing and Liu, Yong and Goh, Rick Siow Mong},
  booktitle={NeurIPS},
  volume={37},
  pages={6625--6647},
  year={2024}
}

@article{van2008visualizing,
  title={Visualizing Data using t-SNE},
  author={van der Maaten, Laurens and Hinton, Geoffrey},
  journal={Journal of Machine Learning Research},
  volume={9},
  pages={2579--2605},
  year={2008}
}

@inproceedings{huang-kiut,
   author = {Huang, Z. and Zhang, X. and Zhang, S.},
   title = {KiUT: Knowledge-injected U-Transformer for Radiology Report Generation},
   booktitle = {CVPR},
   pages = {19809-19818},
   DOI = {10.1109/CVPR52729.2023.01897},
   year = {2023},
}

@inproceedings{liu2024in-context-acmmm,
  title={In-context learning for zero-shot medical report generation},
  author={Liu, Rui and Li, Mingjie and Zhao, Shen and Chen, Ling and Chang, Xiaojun and Yao, Lina},
  booktitle={ACM MM},
  pages={8721--8730},
  year={2024}
}

@inproceedings{irvin-chexpert,
  title={Chexpert: A large chest radiograph dataset with uncertainty labels and expert comparison},
  author={Irvin, Jeremy and Rajpurkar, Pranav and Ko, Michael and Yu, Yifan and Ciurea-Ilcus, Silviana and Chute, Chris and Marklund, Henrik and Haghgoo, Behzad and Ball, Robyn and Shpanskaya, Katie and Seekins, Jayne and Mong, David A. and Halabi, Safwan S. and Sandberg, Jesse K. and Jones, Ricky and Larson, David B. and Langlotz, Curtis P. and Patel, Bhavik N. and Lungren, Matthew P. and Ng, Andrew Y.},
  booktitle={AAAI},
  volume={33},
  number={01},
  pages={590-597},
  DOI={10.1609/aaai.v33i01.3301590},
  year={2019}
}

@inproceedings{jain-radgraph,
   author = {Jain, Saahil and Agrawal, Ashwin and Saporta, Adriel and Truong, Steven and Duong, Du Nguyen Duong Nguyen and Bui, Tan and Chambon, Pierre and Zhang, Yuhao and Lungren, Matthew and Ng, Andrew and Langlotz, Curtis and Rajpurkar, Pranav and Rajpurkar, Pranav},
   title = {Radgraph: Extracting clinical entities and relations from radiology reports},
   pages = {},
   volume = {1},
   booktitle = {NeurIPS},
   year = {2021},
}

@misc{johnson-mimic-cxr-jpg,
      title={MIMIC-CXR-JPG, a large publicly available database of labeled chest radiographs}, 
      author={Alistair E. W. Johnson and Tom J. Pollard and Nathaniel R. Greenbaum and Matthew P. Lungren and Chih-ying Deng and Yifan Peng and Zhiyong Lu and Roger G. Mark and Seth J. Berkowitz and Steven Horng},
      year={2019},
      eprint={1901.07042},
      archivePrefix={arXiv},
      primaryClass={cs.CV},
}

@InProceedings{2019-cvpr-class-balance-loss,
    author = {Cui, Yin and Jia, Menglin and Lin, Tsung-Yi and Song, Yang and Belongie, Serge},
    title = {Class-Balanced Loss Based on Effective Number of Samples},
    booktitle = {CVPR},
    pages={9268--9277},
    month = {June},
    year = {2019}
}

@InProceedings{2017-iccv-focal-loss,
  title={Focal loss for dense object detection},
  author={Lin, Tsung-Yi and Goyal, Priya and Girshick, Ross and He, Kaiming and Doll{\'a}r, Piotr},
  booktitle={ICCV},
  pages={2980--2988},
  year={2017}
}

@misc{oord-cpc-infonce,
      title={Representation Learning with Contrastive Predictive Coding}, 
      author={Aaron van den Oord and Yazhe Li and Oriol Vinyals},
      year={2019},
      eprint={1807.03748},
      archivePrefix={arXiv},
      primaryClass={cs.LG},
}

@inproceedings{radford-learning-clip,
   author = {Radford, Alec and Kim, Jong Wook and Hallacy, Chris and Ramesh, Aditya and Goh, Gabriel and Agarwal, Sandhini and Sastry, Girish and Askell, Amanda and Mishkin, Pamela and Clark, Jack},
   title = {Learning transferable visual models from natural language supervision},
   booktitle = {ICML},
   pages = {8748-8763},
   year = {2021},
}

@inproceedings{wang-mgca,
   author = {Wang, Fuying and Zhou, Yuyin and Wang, Shujun and Vardhanabhuti, Varut and Yu, Lequan},
   title = {Multi-granularity cross-modal alignment for generalized medical visual representation learning},
   booktitle = {NeurIPS},
   volume = {35},
   pages = {33536-33549},
   year = {2022},
}

@inproceedings{pmlr-blip,
  title = 	 {{BLIP}: Bootstrapping Language-Image Pre-training for Unified Vision-Language Understanding and Generation},
  author =       {Li, Junnan and Li, Dongxu and Xiong, Caiming and Hoi, Steven},
  booktitle = 	 {ICML},
  pages = 	 {12888-12900},
  year = 	 {2022},
  volume = 	 {162},
  month = 	 {17-23 Jul},
}

@inproceedings{li2023-blip2,
  title={Blip-2: Bootstrapping language-image pre-training with frozen image encoders and large language models},
  author={Li, Junnan and Li, Dongxu and Savarese, Silvio and Hoi, Steven},
  booktitle={ICML},
  pages={19730--19742},
  year={2023},
  organization={PMLR}
}

@InProceedings{2025-ORID,
    author    = {Gu, Tiancheng and Yang, Kaicheng and An, Xiang and Feng, Ziyong and Liu, Dongnan and Cai, Weidong},
    title     = {ORID: Organ-Regional Information Driven Framework for Radiology Report Generation},
    booktitle = {WACV},
    month     = {February},
    year      = {2025},
    pages     = {378-387}
}

@inproceedings{midl-2025-radialog,
    title={RaDialog: Large Vision-Language Models for X-Ray Reporting and Dialog-Driven Assistance},
    author={Chantal Pellegrini and Ege {\"O}zsoy and Benjamin Busam and Benedikt Wiestler and Nassir Navab and Matthias Keicher},
    booktitle={MIDL},
    year={2025},
    url={https://openreview.net/forum?id=trUvr1gSNI}
}

@misc{chexpert-plus-2024,
      title={CheXpert Plus: Augmenting a Large Chest X-ray Dataset with Text Radiology Reports, Patient Demographics and Additional Image Formats}, 
      author={Pierre Chambon and Jean-Benoit Delbrouck and Thomas Sounack and Shih-Cheng Huang and Zhihong Chen and Maya Varma and Steven QH Truong and Chu The Chuong and Curtis P. Langlotz},
      year={2024},
      eprint={2405.19538},
      archivePrefix={arXiv},
      primaryClass={cs.CL},
}

@article{zhou2022-ijcv-coop,
  title={Learning to Prompt for Vision-Language Models},
  author={Zhou, Kaiyang and Yang, Jingkang and Loy, Chen Change and Liu, Ziwei},
  journal={International Journal of Computer Vision},
  volume={130},
  number={9},
  pages={2337--2348},
  year={2022},
  publisher={Springer US New York}
}

@inproceedings{bao2022beit,
    title={{BE}iT: {BERT} Pre-Training of Image Transformers},
    author={Hangbo Bao and Li Dong and Songhao Piao and Furu Wei},
    booktitle={ICLR},
    year={2022},
    url={https://openreview.net/forum?id=p-BhZSz59o4}
}

@misc{2025-DDaTR-longitudinal,
    author = {{Song}, Shanshan and {Tang}, Hui and {Yang}, Honglong and {Li}, Xiaomeng},
    title = "DDaTR: Dynamic Difference-aware Temporal Residual Network for Longitudinal Radiology Report Generation",
    year = 2025,
    eprint = {2505.03401},
    doi = {10.48550/arXiv.2505.03401},
    archivePrefix = {arXiv},
    primaryClass = {cs.CV},
}

@inproceedings{2017-grad-cam,
  title={Grad-cam: Visual explanations from deep networks via gradient-based localization},
  author={Selvaraju, Ramprasaath R and Cogswell, Michael and Das, Abhishek and Vedantam, Ramakrishna and Parikh, Devi and Batra, Dhruv},
  booktitle={ICCV},
  pages={618--626},
  year={2017}
}

@misc{zhou2025reviewlongitudinalradiologyreport,
      title={A Review of Longitudinal Radiology Report Generation: Dataset Composition, Methods, and Performance Evaluation}, 
      author={Shaoyang Zhou and Yingshu Li and Yunyi Liu and Lingqiao Liu and Lei Wang and Luping Zhou},
      year={2025},
      eprint={2510.12444},
      archivePrefix={arXiv},
      primaryClass={cs.CV},
}

@inproceedings{nips-2023-lvm-med,
     author = {M. H. Nguyen, Duy and Nguyen, Hoang and Diep, Nghiem and Pham, Tan Ngoc and Cao, Tri and Nguyen, Binh and Swoboda, Paul and Ho, Nhat and Albarqouni, Shadi and Xie, Pengtao and Sonntag, Daniel and Niepert, Mathias},
     booktitle = {NeurIPS},
     pages = {27922--27950},
     publisher = {Curran Associates, Inc.},
     title = {LVM-Med: Learning Large-Scale Self-Supervised Vision Models for Medical Imaging via Second-order Graph Matching},
     volume = {36},
     year = {2023}
}

@inproceedings{chen2021-moco-v3,
  title={An empirical study of training self-supervised vision transformers},
  author={Chen, Xinlei and Xie, Saining and He, Kaiming},
  booktitle={ICCV},
  pages={9640--9649},
  year={2021}
}

@inproceedings{hu2022lora,
    title={Lo{RA}: Low-Rank Adaptation of Large Language Models},
    author={Edward J Hu and yelong shen and Phillip Wallis and Zeyuan Allen-Zhu and Yuanzhi Li and Shean Wang and Lu Wang and Weizhu Chen},
    booktitle={ICLR},
    year={2022},
    url={https://openreview.net/forum?id=nZeVKeeFYf9}
}

@inproceedings{delbrouck-etal-2025-srrg,
    title = "Automated Structured Radiology Report Generation",
    author = "Delbrouck, Jean-Benoit  and
      Xu, Justin  and
      Moll, Johannes  and
      Thomas, Alois  and
      Chen, Zhihong  and
      Ostmeier, Sophie  and
      Azhar, Asfandyar  and
      Li, Kelvin Zhenghao  and
      Johnston, Andrew  and
      Bluethgen, Christian  and
      Reis, Eduardo Pontes  and
      Muneer, Mohamed S  and
      Varma, Maya  and
      Langlotz, Curtis",
    booktitle = "ACL",
    month = jul,
    year = "2025",
    doi = "10.18653/v1/2025.acl-long.1301",
    pages = "26813--26829",
    ISBN = "979-8-89176-251-0",
}

@InProceedings{2024-eccv-hergen,
	author="Wang, Fuying
	and Du, Shenghui
	and Yu, Lequan",
	title="HERGen: Elevating Radiology Report Generation with Longitudinal Data",
	booktitle="ECCV",
	year="2025",
	publisher="Springer Nature Switzerland",
	address="Cham",
	pages="183--200",
	doi="10.1007/978-3-031-73001-6_11"
}

@misc{bannur2024maira2groundedradiologyreport,
      title={MAIRA-2: Grounded Radiology Report Generation}, 
      author={Shruthi Bannur and Kenza Bouzid and Daniel C. Castro and Anton Schwaighofer and Sam Bond-Taylor and Maximilian Ilse and Fernando Pérez-García and Valentina Salvatelli and Harshita Sharma and Felix Meissen and Mercy Ranjit and Shaury Srivastav and Julia Gong and Fabian Falck and Ozan Oktay and Anja Thieme and Matthew P. Lungren and Maria Teodora Wetscherek and Javier Alvarez-Valle and Stephanie L. Hyland},
      year={2024},
      eprint={2406.04449},
      archivePrefix={arXiv},
      primaryClass={cs.CL},
}

@misc{zhang2025biomedclipmultimodalbiomedicalfoundation,
      title={BiomedCLIP: a multimodal biomedical foundation model pretrained from fifteen million scientific image-text pairs}, 
      author={Sheng Zhang and Yanbo Xu and Naoto Usuyama and Hanwen Xu and Jaspreet Bagga and Robert Tinn and Sam Preston and Rajesh Rao and Mu Wei and Naveen Valluri and Cliff Wong and Andrea Tupini and Yu Wang and Matt Mazzola and Swadheen Shukla and Lars Liden and Jianfeng Gao and Angela Crabtree and Brian Piening and Carlo Bifulco and Matthew P. Lungren and Tristan Naumann and Sheng Wang and Hoifung Poon},
      year={2025},
      eprint={2303.00915},
      archivePrefix={arXiv},
      primaryClass={cs.CV},
}

@ARTICLE{zhang-kad,
   author = {Zhang, Xiaoman and Wu, Chaoyi and Zhang, Ya and Xie, Weidi and Wang, Yanfeng},
   title = {Knowledge-enhanced visual-language pre-training on chest radiology images},
   journal = {Nature Communications},
   volume = {14},
   number = {1},
   pages = {4542},
   ISSN = {2041-1723},
   year = {2023},
   DOI  = {10.1038/s41467-023-40260-7},
   type = {Journal Article}
}

@article{ikezogwo2023quilt-1m,
	title={Quilt-1m: One million image-text pairs for histopathology},
	author={Ikezogwo, Wisdom and Seyfioglu, Saygin and Ghezloo, Fatemeh and Geva, Dylan and Sheikh Mohammed, Fatwir and Anand, Pavan Kumar and Krishna, Ranjay and Shapiro, Linda},
	journal={NeurIPS},
	volume={36},
	pages={37995--38017},
	year={2023}
}

@inproceedings{wu-medklip,
    author    = {Wu, Chaoyi and Zhang, Xiaoman and Zhang, Ya and Wang, Yanfeng and Xie, Weidi},
    title     = {MedKLIP: Medical Knowledge Enhanced Language-Image Pre-Training for X-ray Diagnosis},
    booktitle = {ICCV},
    month     = {October},
    year      = {2023},
    pages     = {21372-21383}
}

@inproceedings{wang2023metransformer,
  title={METransformer: Radiology Report Generation by Transformer with Multiple Learnable Expert Tokens},
  author={Wang, Zhanyu and Liu, Lingqiao and Wang, Lei and Zhou, Luping},
  booktitle={CVPR},
  pages={11558-11567},
  doi={10.1109/CVPR52729.2023.01112},
  year={2023}
}

@misc{Sanh2019DistilBERTAD,
      title={DistilBERT, a distilled version of BERT: smaller, faster, cheaper and lighter}, 
      author={Victor Sanh and Lysandre Debut and Julien Chaumond and Thomas Wolf},
      year={2020},
      eprint={1910.01108},
      archivePrefix={arXiv},
      primaryClass={cs.CL},
}

@article{2020-eye-gaze-data,
  title={Eye gaze data for chest x-rays},
  author={Karargyris, Alexandros and Kashyap, Satyananda and Lourentzou, Ismini and Wu, Joy and Tong, Matthew and Sharma, Arjun and Abedin, Shafiq and Beymer, David and Mukherjee, Vandana and Krupinski, Elizabeth and others},
  journal={PhysioNet https://doi. org/10.13026/QFDZ-ZR67},
  year={2020}
}

@inproceedings{2024-ACCV-eye-gaze-fg-cxr,
  title={Fg-cxr: a radiologist-aligned gaze dataset for enhancing interpretability in chest x-ray report generation},
  author={Pham, Trong Thang and Ho, Ngoc-Vuong and Bui, Nhat-Tan and Phan, Thinh and Brijesh, Patel and Adjeroh, Donald and Doretto, Gianfranco and Nguyen, Anh and Wu, Carol C and Nguyen, Hien and others},
  booktitle={ACCV},
  pages={941--958},
  year={2024}
}

@article{2022-reflacx,
   title={REFLACX, a dataset of reports and eye-tracking data for localization of abnormalities in chest x-rays},
   volume={9},
   DOI={10.1038/s41597-022-01441-z},
   number={1},
   journal={Scientific Data},
   publisher={Springer Science and Business Media LLC},
   author={Bigolin Lanfredi, Ricardo and Zhang, Mingyuan and Auffermann, William F. and Chan, Jessica and Duong, Phuong-Anh T. and Srikumar, Vivek and Drew, Trafton and Schroeder, Joyce D. and Tasdizen, Tolga},
   year={2022},
   month=jun
}

@inproceedings{2024-accv-eye-cls,
  title={Seeing Through Expert's Eyes: Leveraging Radiologist Eye Gaze and Speech Report with Graph Neural Networks for Chest X-ray Image Classification},
  author={Sultana, Jamalia and Qin, Ruwen and Yin, Zhaozheng},
  booktitle={ACCV},
  pages={2579--2595},
  year={2024}
}

@inproceedings{2020-iclr-bertscore,
    title={BERTScore: Evaluating Text Generation with BERT},
    author={Tianyi Zhang* and Varsha Kishore* and Felix Wu* and Kilian Q. Weinberger and Yoav Artzi},
    booktitle={ICLR},
    year={2020},
    url={https://openreview.net/forum?id=SkeHuCVFDr}
}

@InProceedings{icml-2021-perceiver,
  title = 	 {Perceiver: General Perception with Iterative Attention},
  author =       {Jaegle, Andrew and Gimeno, Felix and Brock, Andy and Vinyals, Oriol and Zisserman, Andrew and Carreira, Joao},
  booktitle = 	 {ICML},
  pages = 	 {4651--4664},
  year = 	 {2021},
  volume = 	 {139},
  month = 	 {18--24 Jul},
  publisher =    {PMLR},
}

@article{PriorRG,
    title={PriorRG: Prior-Guided Contrastive Pre-training and Coarse-to-Fine Decoding for Chest X-ray Report Generation},
    volume={40},
    DOI={10.1609/aaai.v40i9.37657},
    number={9},
    journal={AAAI},
    author={Liu, Kang and Ma, Zhuoqi and Fang, Zikang and Li, Yunan and Xie, Kun and Miao, Qiguang},
    year={2026},
    month={Mar.},
    pages={7206-7214} 
}

@InProceedings{2024-miccai-eye-vlm,
    author="Kim, Yunsoo
    and Wu, Jinge
    and Abdulle, Yusuf
    and Gao, Yue
    and Wu, Honghan",
    title="Enhancing Human-Computer Interaction in Chest X-Ray Analysis Using Vision and Language Model with Eye Gaze Patterns",
    booktitle="MICCAI",
    year="2024",
    publisher="Springer Nature Switzerland",
    address="Cham",
    pages="184--194",
    doi="10.1007/978-3-031-72384-1\_18"
}

@inproceedings{wang2024-wacv-gazegnn,
  title={Gazegnn: A gaze-guided graph neural network for chest x-ray classification},
  author={Wang, Bin and Pan, Hongyi and Aboah, Armstrong and Zhang, Zheyuan and Keles, Elif and Torigian, Drew and Turkbey, Baris and Krupinski, Elizabeth and Udupa, Jayaram and Bagci, Ulas},
  booktitle={WACV},
  pages={2194--2203},
  year={2024}
}

@inproceedings{acl-eye-2025-look,
    title = "Look {\&} Mark: Leveraging Radiologist Eye Fixations and Bounding boxes in Multimodal Large Language Models for Chest {X}-ray Report Generation",
    author = "Kim, Yunsoo  and
      Wu, Jinge  and
      Kim, Su Hwan  and
      Vasudev, Pardeep  and
      Shen, Jiashu  and
      Wu, Honghan",
    booktitle = "ACL",
    publisher = "Association for Computational Linguistics",
    month = jul,
    year = "2025",
    address = "Vienna, Austria",
    doi = "10.18653/v1/2025.findings-acl.909",
    pages = "17680--17694",
}

@inproceedings{acl-2025-libra,
    title = "Libra: Leveraging Temporal Images for Biomedical Radiology Analysis",
    author = "Zhang, Xi  and
      Meng, Zaiqiao  and
      Lever, Jake  and
      Ho, Edmond S. L.",
    booktitle = "ACL",
    month = jul,
    year = "2025",
    address = "Vienna, Austria",
    publisher = "Association for Computational Linguistics",
    doi = "10.18653/v1/2025.findings-acl.888",
    pages = "17275--17303",
}

@misc{riju2025-arxiv-eye,
  title={Eyes on the Image: Gaze Supervised Multimodal Learning for Chest X-ray Diagnosis and Report Generation},
  author={Riju, Tanjim Islam and Anwar, Shuchismita and Joy, Saman Sarker and Sadeque, Farig and Shatabda, Swakkhar},
  journal={arXiv preprint arXiv:2508.13068},
  year={2025}
}

@article{2024-nejm-ai-med-palm,
  title={Towards generalist biomedical AI},
  author={Tu, Tao and Azizi, Shekoofeh and Driess, Danny and Schaekermann, Mike and Amin, Mohamed and Chang, Pi-Chuan and Carroll, Andrew and Lau, Charles and Tanno, Ryutaro and Ktena, Ira and others},
  journal={NEJM AI},
  volume={1},
  number={3},
  pages={AIoa2300138},
  year={2024},
  publisher={Massachusetts Medical Society}
}

@article{ma2024-nips-eye-egma,
  title={Eye-gaze guided multi-modal alignment for medical representation learning},
  author={Ma, Chong and Jiang, Hanqi and Chen, Wenting and Li, Yiwei and Wu, Zihao and Yu, Xiaowei and Liu, Zhengliang and Guo, Lei and Zhu, Dajiang and Zhang, Tuo and others},
  journal={NeurIPS},
  volume={37},
  pages={6126--6153},
  year={2024}
}

@inproceedings{ml4h-indication-rg,
  title = 	 {Pragmatic Radiology Report Generation},
  author =       {Nguyen, Dang and Chen, Chacha and He, He and Tan, Chenhao},
  booktitle =  {ML4H},
  publisher =    {PMLR},
  pages = 	 {385-402},
  year = 	 {2023},
  volume = 	 {225},
}

@article{2026-nbe-afloc,
  title={A multimodal vision--language model for generalizable annotation-free pathology localization},
  author={Yang, Hao and Zhou, Hong-Yu and Liu, Jiarun and Huang, Weijian and Li, Cheng and Li, Zhihuan and Gao, Yuanxu and Liu, Qiegen and Liang, Yong and Yang, Qi and others},
  journal={Nature Biomedical Engineering},
  pages={1--15},
  year={2026},
  doi={10.1038/s41551-025-01574-7},
  publisher={Nature Publishing Group UK London}
}

\end{document}